\pdfoutput=1

\documentclass[11pt]{article}

\usepackage[preprint]{acl}

\usepackage{times}
\usepackage{latexsym}
\usepackage{amsmath}
\usepackage{booktabs}
\usepackage{xcolor}
\usepackage{listings}
\usepackage{mdframed}

\usepackage[T1]{fontenc}

\usepackage[utf8]{inputenc}

\usepackage{microtype}

\usepackage{inconsolata}

\usepackage{graphicx}

\title{Applying Large Language Models to Characterize Public Narratives}

\author{Elinor Poole-Dayan\thanks{~~Equal contribution} \\
  MIT\\\texttt{elinorpd@mit.edu} \\ \And Daniel T. Kessler\textsuperscript{$\ast$}\\MIT\\\texttt{kessler1@mit.edu} \\\And  Hannah Chiou\\Wellesley College\\\AND  Margaret Hughes\\MIT
   \\\And Emily S. Lin \\Harvard University \\\And 
  Marshall Ganz \\
  Harvard University \\
  \And Deb Roy \\ MIT}

\begin{document}
\maketitle
\begin{abstract}
Public Narratives (PNs) are key tools for leadership development and civic mobilization, yet their systematic analysis remains challenging due to their subjective interpretation and the high cost of expert annotation. In this work, we propose a novel computational framework that leverages large language models (LLMs) to automate the qualitative annotation of public narratives. Using a codebook we co-developed with subject-matter experts, we evaluate LLM performance against that of expert annotators. 
Our work reveals that LLMs can achieve near-human-expert performance, achieving an average F1 score of 0.80 across 8 narratives and 14 codes. We then extend our analysis to empirically explore how PN framework elements manifest across a larger dataset of 22 stories. Lastly, we extrapolate our analysis to a set of political speeches, establishing a novel lens in which to analyze political rhetoric in civic spaces.
This study demonstrates the potential of LLM-assisted annotation for scalable narrative analysis and highlights key limitations and directions for future research in computational civic storytelling.
\end{abstract}

\begin{figure*}
    \centering
    \includegraphics[width=0.9\linewidth]{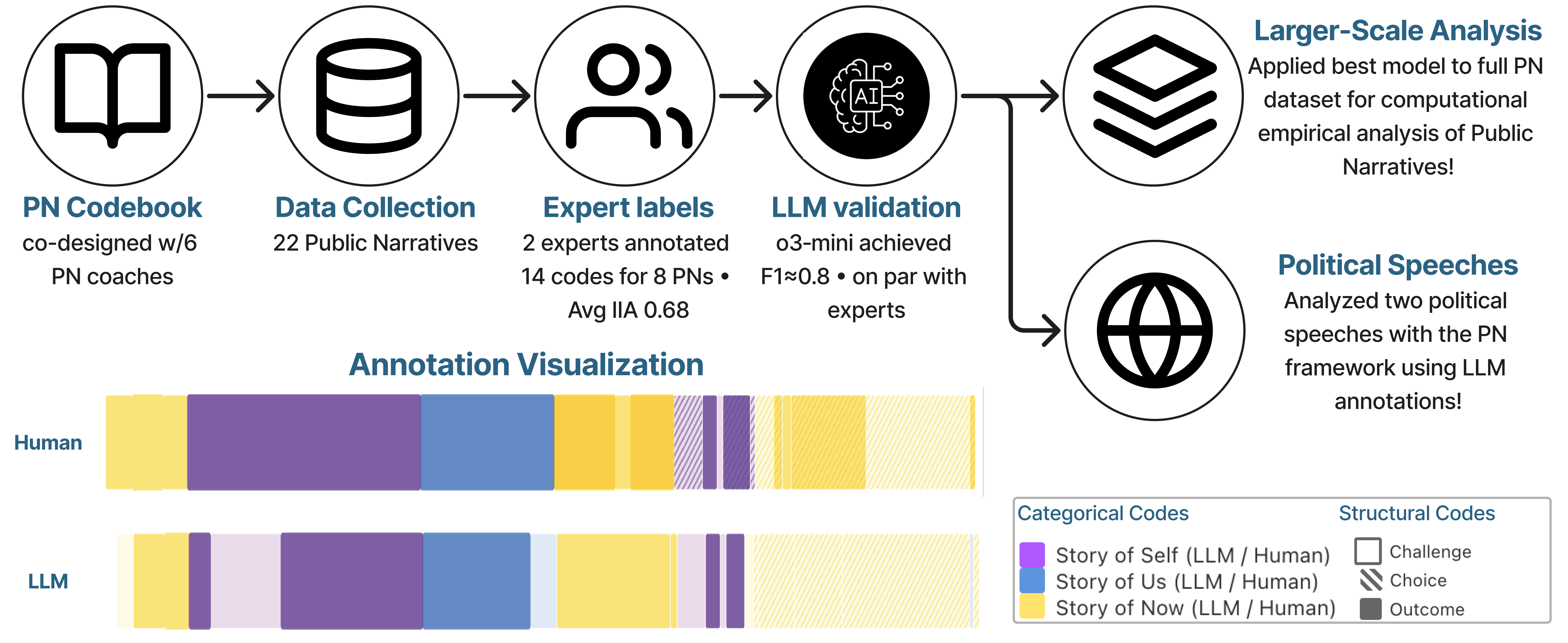}
    \caption{We compare an LLMs' ability to annotate Public Narratives to human experts following our codebook co-developed with experts. To visualize the annotations,  the length of each portion corresponds to the number of sentences coded. }
    \label{fig:main}
\end{figure*}

\section{Introduction}
While narratives have been used within politics and leadership domains for centuries—to illustrate issue complexity and urgency, causality \citep{hampton_enhancing_2004}, and possible futures \citep{oneill_roads_2017}, with a central aim of motivating collective action—only recently have pedagogical approaches been developed to train civic actors in their contextually-specific use. One such approach is the Public Narrative framework (PN), a unique leadership development practice that teaches narrative framing to motivate collective action \cite{ganz_crafting_2023}. Between 2006 and 2016 alone, at least 32,184 people participated in 448 workshops across 25 countries, applying PN in fields such as education, health care, advocacy, and politics \cite{aiello_public_2020}. Notably, PN was a foundational element in training volunteer leadership teams for Barack Obama's 2008 campaign, which ultimately mobilized 2.2 million volunteers—more than eight times the comparable number in the 2004 U.S. election \cite{mckenna_groundbreakers_2015}. 

However, despite broad impact and adoption, and beyond the relatively few experts who currently teach it, PN remains challenging to precisely define and analyze. Many of its key features—values-based leadership, moral agency, and the strategic use of urgency and hope—lack clear empirical methods for their systematic evaluation, and may even be interpreted differently by different subject experts trained in their application. While forthcoming research \citep{lerner2025psychology} proposes psychological dimensions of a persuasive PN to correlate with impacts with real-world behaviors, no formal computational approach has been developed to analyze how existing or "in-the-wild" narratives (e.g., political speeches) align with the PN framework, such that systematic analysis and cross-comparison of their feature-specific impacts across civic spaces becomes possible. Consequently, the human text annotations that are required for such work are time-consuming, costly, and difficult to scale. If we seek to analyze large datasets of PNs, a computational solution is necessary. 

Given recent advances in large language models (LLMs) and their effectiveness in structured content analysis \citep{ziems-etal-2024-large, ruckdeschel-2025-just, xiao-qual-2023}, we propose LLMs as a viable tool for large-scale PN annotation and test their performance against PN expert annotators. This work makes four key contributions: 

\begin{enumerate}
    \item A formalized PN annotation codebook (an instructive rubric for qualitative annotation of text features), co-developed with six subject experts, enabling structured human annotation and automated analysis of PN;
    \item An empirical validation of LLMs as a scalable PN annotation method on 8 stories, achieving 0.80 average F1 in replicating human expert annotations; 
    \item A discussion relating our results to prior work and takeaways for leveraging LLMs as qualitative annotators to guide future work; and
    \item An effective application  to two political speeches, demonstrating wider applications of our approach to computational social science and civic spaces.
\end{enumerate}

Our work is the first to use computational methods to automate understanding of PNs. Here, we provide empirical findings to lay groundwork for future large-scale evaluations of PNs, opening new possibilities for understanding civic engagement and leadership development as well as the real-world impacts that unique features of narration have in social spaces.

\section{Related Works}

\subsection{Understanding Civic Narratives}
Narrative is a historically prevalent and impactful medium for civic organizing and participation \citep{jenkins_path_2024}. The unique affordances of narrative framing—emotional activation \cite{bilandzic_narrative_2019}, issue comprehension \citep{zwaan_five_1999}, trust formation \cite{clementson_narrative_2020}, deictic shift \cite{appel_transportation_2010}, and persuasion \cite{hamby_happily_2016}, among others—make it well-suited for contexts that seek to leverage individual and collective civic identities \cite{adler_living_2012} \citep{haste_power_2017}) to motivate real-world action \citep{dimond_hollaback_2013}. There are many conceptual frameworks that scientists and practitioners may use to understand such narration practices—from the Jungian-inspired Hero's Journey \cite{campbell_hero_2008} to the more contemporary Narrative Policy and Engagement Frameworks \citep{miller-day_narrative_2013,shanahan_narrative_2018,bilandzic_narrative_2019}. However, narratives are structured in many ways, composed in each case to suit particular functional purposes \cite{bamberg_why_2012}, and their method of analysis should therefore be suitable to the context in which the narrative was developed and the function for which it was created. For example, Labovian oral narrative may be a suitable framework for examining stream-of-consciousness spoken narrations, but not for understanding text-based digital media. By the same token, while much is known about how narratives are used and transmitted in civic organizing contexts, few frameworks are appropriately positioned to develop or understand them. PNs uniquely excel in such framing, but they remain poorly understood from a systematic and empirical perspective.

\subsection{LLMs for Qualitative Analysis}
Recent advances in NLP have enabled more widespread use of LLMs for qualitative textual analysis and annotation tasks. In particular, leveraging LLMs as zero or few-shot  annotators has been shown to be extremely promising \citep{gilardi-chatgpt-2023, wang-etal-2021-want-reduce, ding-etal-2023-gpt, he-etal-2024-annollm, huang-chatgpt-2023, ziems-etal-2024-large}. LLM-assisted content analysis (LACA) has shown effectiveness in reducing the time burden of deductive coding while maintaining human-level accuracy for well-defined coding schemes \citep{chew_llm-assisted_2023}, potentially even for subjective, nuanced tasks using codebooks \citep{ruckdeschel-2025-just,lupo_towards_2024,xiao-qual-2023}. While human validation of LLM performance remains paramount \citep{pangakis2023automated}, this line of work may open up NLP research to tackle more complex, interdisciplinary, or niche datasets for which human annotation is very difficult or expensive \citep{ruckdeschel-2025-just}.

However, other prior works have found LLMs struggle with more complex, subjective or context-dependent tasks in NLP tasks such as annotating code or sentiment analysis \citep{ahmed_can_2024,li-etal-2023-synthetic}. \citet{ashwin_using_2023} raise concerns of bias\footnote{This refers to bias in the technical sense, "that the errors that LLMs make in annotating interview transcripts are not random with respect to the characteristics of the interview subjects" \citep{ashwin_using_2023}.} in LLM-based qualitative annotation. Likewise, in automated story analysis using GPT-3.5 and Llama 2, \citet{chhun_language_2024} found that 31\% of LLM-generated annotation explanations lacked direct references to the story being evaluated. \citet{chen-si-2024-reflections} proposed a dual-agent model for automated story annotation, finding that their
system performed well in identifying basic structural narrative elements (e.g., plot points), but failed at accounting for ambiguity or nuance (e.g., thematic depth, possible alternative interpretations). Notably, the LLM stayed true to the coding scheme provided, although the scheme was developed using another LLM instance. 
In the case of automatic grading using Mixtral-8x7b, \citet{wu_unveiling_2024} find an alignment gap between human and LLM scores, partially due to the model taking heuristic shortcuts, skipping deeper logical reasoning. On the other hand, they find that high-quality analytic rubrics can improve LLM accuracy and argue that aligning LLM outputs to human expectations is important for reducing the performance gap between LLMs and humans. 

In a setup more similar to ours, \citet{xiao-qual-2023} find GPT-3 has "fair to substantial" agreement with human experts in implementing a codebook. Additionally, they find that prompts centered around the codebook were more effective than those with examples. \citet{lupo_towards_2024} similarly use LLMs to annotate public policy documents and find GPT-4 models matched or outperformed humans even the on subjective annotation tasks when given a detailed codebook.

\subsection{Our Work}

We build upon prior research by developing an annotation scheme for PNs (Section~\ref{codebook}), then systematically testing whether LLMs can annotate these narratives as effectively as human experts (Section~\ref{eval}). Unlike previous work focused on general content analysis \citep{chew_llm-assisted_2023} or automated grading \citep{wu_unveiling_2024,xiao_human-ai_2024}, we focus on narratives used in politics, community organizing, and movement building—a domain where story effectiveness is tied to rhetorical structure, emotional engagement, and audience resonance. Based on prior findings and recommendations for best practices leveraging LLMs for subjective, context-dependent text annotation tasks, we design our LLM prompts to be centered around our expert-validated codebook to increase alignment between LLM and human annotations \citep{xiao-qual-2023,ruckdeschel-2025-just,wu_unveiling_2024,lupo_towards_2024,tornberg_best_2024}. Moreover, we use much more recent, highly capable reasoning models such as OpenAI's \texttt{o3-mini} \citep{openai_o3mini_2025}. By applying structured annotation schemes and evaluating LLM performance, we contribute to both computational narrative understanding and the broader discourse on AI-assisted qualitative research.

\section{Defining Public Narrative} 
Before constructing a codebook for LLM-human annotation comparison, we first formalize our conception of Public Narratives. PNs are developed to harness storytelling to communicate values, enable agency, and inspire action, emphasizing that effective leadership engages the "head," "heart," and "hands": aligning strategy, motivation, and action \cite{ganz_what_2009}. Narrative is a particularly helpful tool in leadership, and is often most needed under conditions of uncertainty, when collective sensemaking is necessary to enable people to move toward shared purpose \citep{ganz_people_2024}. Toward these ends, a Public Narrative itself is a unique artifact of a coached narration process, composed of three linked elements: Story of Self (SoS), Story of Us (SoU), and Story of Now (SoN). SoS conveys a leader's origin story—the moment or experience that shaped their core values and commitment to action. Often rooted in formative life experiences, these stories also reveal sources of hope that drive their continued commitment to their cause. The SoU illustrates how a community or group embodies shared values through collective experience and action, fostering a sense of belonging, or "us-ness," that strengthens solidarity and motivation. Finally, SoN establishes urgency and calls the audience—the "us"—to action. It presents a vision of what could be achieved if action is taken (the dream) and contrasts it with the likely consequences of inaction (the nightmare). While each of these stories can be used in various contexts, to bring them together into one PN is referred to as a "Linked Story," which is the primary story structure we work with in this study.

Each story within the PN framework follows a fundamental structure (Challenge → Choice → Outcome) that seeks to ground abstract personal values within concrete experiences, illustrating a moment of adversity or uncertainty and using the responsive choice to showcase valued action to address a collective challenge. Here, an outcome demonstrates the result of that choice, revealing the stakes and consequences of action (or inaction). Each of the three linked stories shares a unique relationship with these structural features. For a SoS, this structure often unfolds through formative moments from one's youth or early leadership experiences, whereas a SoU highlights collective challenges and a shared decision to take action towards a collective outcome, and a SoN underscores the present crisis and serves as a call-to-action for audiences. 
Beyond structure, there are many other important features are taught to make PNs both compelling and influential. Additional information is in Appendix~\ref{appendix:pn}.

\section{Codebook Development}\label{codebook} 
In qualitative annotation tasks, human annotators read pieces of text line-by-line alongside a rubric ("codebook") to isolate where different features appear in the text. Consequently, effective qualitative annotation relies on the construction of clear, substantive definitions of systematic features—called codes—to be identified or described \citep{williams_art_2019}. In this case, the features are those that constitute the core elements of PNs. But what if the required feature gives rise cognitively—within or even across human annotators—to the feature sought rather than embodied within the content being analyzed (e.g., a feeling of hope evoked in the annotator, rather than one expressed intentionally by the author)? Being a framework geared toward collective mobility, PNs utilize many formal concepts whose essential nature lies more in the impact elicited (e.g., a feeling of hope or urgency) than in the content that gives rise to it. Over a period of months or years, PN coaches learn to identify the elements of narration that inspire audiences to access a sense of hope in collective action, shared values, and urgency, for example. While such content may take a certain thematic shape, it often possesses extreme variability across storytellers, often making the unambiguous codebook-guided labeling of such content subjective. To account for this diversity of human interpretation, we draw from recent work exploring the ways in which LLMs may be used, rather than to "wash over" human annotator disagreement, to account for human subjectivity by allowing for multiple possible interpretations of a particular piece of text \citep{plank_problem_2022}.

To develop an initial qualitative codebook 
suitable for use by both humans and LLMs, we develop and test a coding scheme drawing upon the materials (e.g., worksheets, coaching manuals) used to instruct PN coaches and students. We iteratively test this codebook (83 codes across seven coding categories) on two publicly available gold-standard sample narratives used commonly for instruction. 
To finalize our codebook, we then recruit six professional PN coaches to annotate these two narratives using iterative versions of our codebook, with each annotation task followed by a qualitative interview for feedback. After five codebook iterations and six participant interviews, we finalize a codebook of 14 codes across three coding types (categorical, structural, and content codes). 

\section{Public Narrative Data Collection}\label{datacollection}
To evaluate LLM annotation performance on a larger dataset, we collected PNs from former participants (spanning $\approx$5 years) in Marshall Ganz's \textit{Leadership, Organizing, and Action} (LOA) course at the Harvard Kennedy School. These narratives were originally developed throughout the course and recorded as part of the course's final module. Through an IRB-approved protocol, we obtained consent\footnote{As per the IRB protocol, participants were contacted by Harvard personnel and provided informed consent via DocuSign to share their stories for research purposes.} from 22 participants to analyze their recorded PNs. 

Following consent, we transcribed the recordings using AssemblyAI's speech-to-text API.\footnote{To protect privacy, we kept only the audio of each narrative; transcripts were edited only to remove speaker introductions and occasional background audience speech to remove any identifying information.} We then applied NLTK's sentence tokenizer to segment each narrative by sentence for annotation. 

All 22 transcripts were annotated using our LLM schema; descriptive statistics of the 22 PNs can be found in Appendix Table~\ref{appendix:desc-stats}. Of 22 total transcripts, we randomly selected eight\footnote{This number was determined by the annotators' availabilities and time constraints.} for annotation by two experts. Using the final codebook definitions, the annotators applied all 14 binary codes to each sentence of the selected narratives. Each sentence was annotated using a binary scheme (1 = present, 0 = absent). Annotations were conducted in a spreadsheet interface, where all codes and their definitions were viewed simultaneously. Although no formal annotation order was required, annotators typically passed through each narrative three times—first for categorical codes (SoS, SoU, SoN), next for structural codes (Challenge, Choice, Outcome), and finally for the remaining content codes. Each story took approximately 20–30 minutes per annotator. To ensure consistency, annotators followed the definitions and usage guidelines in the codebook, which specified exactly how and when to apply each code.
\begin{figure*}[t]
    \centering
    \includegraphics[width=\linewidth]{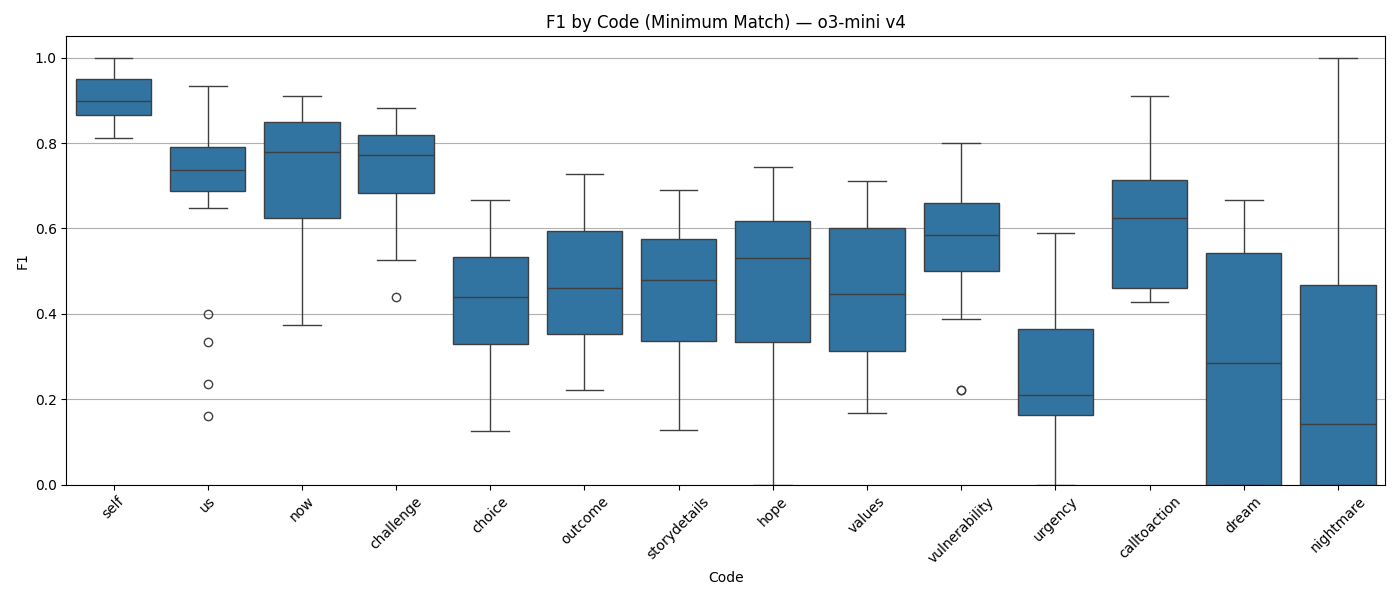}
    \caption{The average F1 scores per code of \texttt{o3-mini} with \textsc{CoT + Prompt Chaining} compared to Minimum Match human annotations across the eight annotated stories. The bars represent the standard deviation across 3 runs.
    }
    \label{f1boxplot}
\end{figure*} 
\subsection{Human Inter-Annotator Agreement (IIA)}\label{ann-agreement}

To assess the reliability of our codebook and establish a human benchmark for LLM evaluation, we measure IIA across the eight doubly-annotated PN described above. 
We report three agreement metrics: raw percent agreement ($p_o$), Cohen's $\kappa$, and the prevalence- and bias-adjusted kappa (PABAK). As Table~\ref{tab:agreement} shows, agreement across codes is high ($p_o = 0.84$ on average). PABAK provides a more robust measure of agreement in cases of class imbalance ($0.68$ average), particularly where $\kappa$ underrepresents agreement on absent codes (global $\kappa = 0.49$). Thresholds for acceptable reliability vary across domains \cite{mchugh_interrater_2012}; our results meet or exceed typical standards for subjective narrative annotation. Still, variation in agreement across codes suggests that some PN features remain inherently interpretive. Further discussion is available in Appendix~\ref{appendix:annotator-agreement}.

\section{LLM Methodology} \label{llm-experiments}

In order to better understand PNs through this framework, there is a need for a larger scale dataset of annotated narrative samples. Not only is this a difficult task even for experts, it is very time consuming and therefore expensive to collect such a labeled dataset, and manual analysis thereof would be infeasible. Given the recent rise in capabilities exhibited by LLMs and their increasing use in complex qualitative text analysis tasks, we test whether state-of-the-art LLMs have the potential to achieve near-human performance in annotating PNs. To this end, we conduct several prompting experiments across a few highly capable LLMs and analyze their efficacy in annotating eight PNs against the expert annotations we collected above. 

\subsection{Prompt and Model Selection}
To identify the best model and prompting technique, we perform preliminary experiments on a subset of the six core codes (3 categorical and 3 structural codes) on the two exemplar narratives annotated each by 3 PN experts (Section~\ref{codebook}). We experiment with both \texttt{gpt-4o-mini} and \texttt{o3-mini} \citep{openai_gpt4o_2024,openai_o3mini_2025}, testing \texttt{gpt-4o-mini} using two different output modes available via the API: structured and predicted outputs.\footnote{OpenAI documentation: \href{https://platform.openai.com/docs/guides/structured-outputs}{structured} \& \href{https://platform.openai.com/docs/guides/predicted-outputs}{predicted}. Currently, \texttt{o3-mini} only supports structured outputs.} This gives us a total of 3 models to compare between.
We test three prompting techniques described in Appendix~\ref{appendix:prompting} and full prompts are in Appendix~\ref{appendix:prompts}.

Overall, the best performing model-prompt configuration was \texttt{o3-mini} and \textsc{CoT + Prompt Chaining}. Full results are in Appendix~\ref{appendix:more-results}. 

\section{Evaluating LLM-Human Performance}\label{eval}
To evaluate the annotations of the best performing model--\texttt{o3-mini} and \textsc{CoT + Prompt Chaining}--we use this setup to annotate the eight PNs collected in Section~\ref{datacollection}, each of which were annotated by two experts. We evaluate performance on all 14 codes: categorical codes, \textit{(SoS, SoU, SoN)}; structural codes \textit{(Challenge, Choice, Outcome)}; and content codes \textit{(Story Details, Hope, Values, Vulnerability, Urgency, Call-to-Action, Dream, and Nightmare)}.\footnote{We extend the prompt schema with a 3rd prompt chain step to annotate the content codes. The complete final prompts with codebook definitions are in Appendix~\ref{appendix:finalprompt}.}

To compare LLM annotation performance to humans, we report the weighted F1 score across the 14 codes. While in other domains one often takes the majority annotation as the gold label, in this context, if at least one expert annotated the presence of a code under their interpretation, this is considered valid \citep{plank_problem_2022}. Thus, for each experiment, we compare both against the Majority human annotation as well as the Minimum Match, where the LLM must match at least one annotator \citep{piper-bagga-2024-using}. For all experiments, we perform three runs.

\subsection{Results} \label{sec:results}
We observe that \texttt{o3-mini} achieves impressive performance across all codes, with Micro-/Macro-F1 scores of 0.87/0.75 compared to the Majority human annotations and 0.82/0.75 Micro-/Macro-F1 compared to the Minimum Match human annotations. Looking into the results by individual codes (Figure~\ref{f1boxplot}), we see that the model consistently excels on the categorical codes (SoS, SoU, SoN) as well as challenge and call to action. On the other hand, we observe decreased performance on the dream, nightmare, and urgency codes.

\section{Extrapolation to a Larger PN Dataset}
Our results demonstrate that \texttt{o3-mini} achieves near-human performance in applying our PN codebook, enabling large-scale, automated analysis of Public Narratives. Using the same model and prompt, we annotate all 22 stories with all 14 codes, conducting three runs per story and taking the majority vote across model outputs. This annotated dataset allows us to explore patterns in (1) code frequency and distribution across stories (\ref{freq}), (2) code co-occurrence (\ref{jaccard}), and (3) pairwise code correlations (below), which we interpret in light of expectations from Public Narrative theory.

\subsection{Code Correlations}\label{pearson}
To better understand how codes may appear in PNs in relation to each other, we compute pairwise Pearson correlations across all 14 codes for the 22 LLM-annotated transcripts (see Figure~\ref{pearson-heatmap}). The strongest positive correlation emerged between SoN and call to action ($r = 0.513$), which aligns closely with theoretical expectations — concrete calls to action often follow the establishment of immediate stakes or temporal urgency. Choice and call to action ($r = 0.468$) were highly correlated, suggesting that model annotations identified moments of decision as narrative pivot points leading to action. Together, these codes make intuitive sense and point to the common PN usage of temporal framing to create tension and motivate listeners to action. The correlation between outcome and hope ($r = 0.404$) also matched intuition, as many stories or anecdotes may resolve themselves with a future-looking sense of possibility. On the other hand, SoS and SoN were strongly negatively associated ($r = -0.638$), suggesting that these codes describe different narrative moments in the context of a PN and are used for different purposes. Similarly, SoS and SoU were negatively correlated ($r = -0.486$). This moderate negative association suggests that speakers tend to alternate between highlighting individual experience (self) and invoking collective identity (us), rather than blending them within the same narrative unit. The strong negative correlations among the three categorical codes suggest that there is a rhetorical separation and flow in PNs, in which a storyteller may first ground an issue in individual stakes before pivoting toward collective and temporal stakes.

\section{Extrapolation to Political Speeches}
\begin{figure*}
    \centering
    \includegraphics[width=1\linewidth]{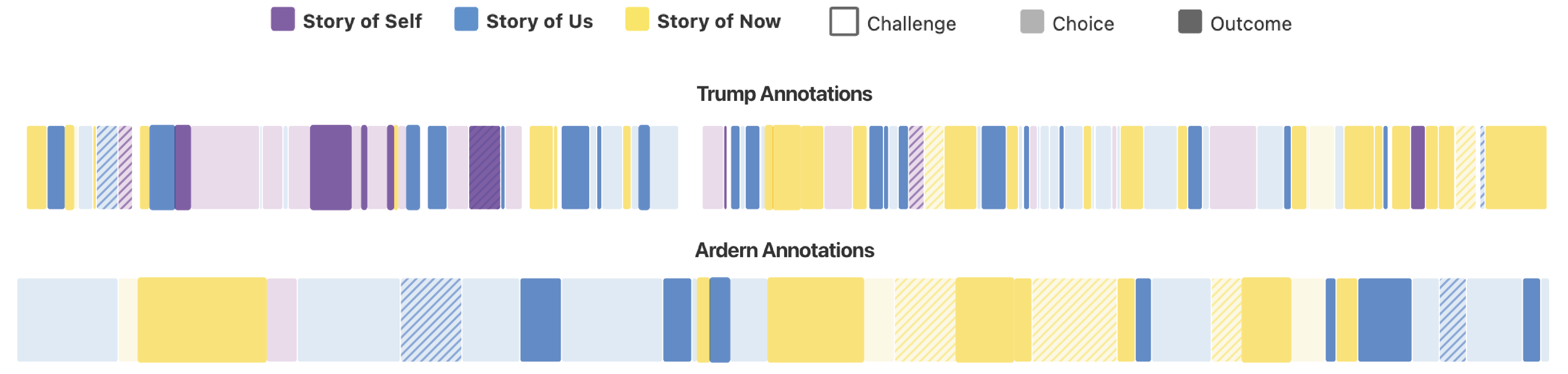}
    \caption{Comparison of a Donald Trump and Jacinda Ardern political speech automatically annotated using \texttt{o3-mini} with our PN codebook.}
    \label{fig:trump-Ardern}
\end{figure*}

Public Narrative both informs civic leaders' speeches, and is informed by how civic leaders have given effective speeches. Therefore, while politicians may not leverage Public Narrative explicitly in their speeches, we would expect to be able to recognize PN components in "in-the-wild" political speeches, and differences across politicians with varying messages and styles. To investigate the further use of our approach for political speeches, we annotate two recent political speeches using \texttt{o3-mini}.  For this comparison, each speech had a unique context that we hoped to explore by way of \textit{which} features of Public Narratives presented more or less prominently. The first was Donald Trump's RNC acceptance speech following a recent assassination attempt at his public rally in July 2024. The second speech was New Zealand Prime Minister Jacinda Ardern's public speech following the Christchurch mosque shooting. We analyze speeches both quantitatively (using our existing framework) and qualitatively.

In Figure~\ref{fig:trump-Ardern}, we see that Ardern's story is primarily comprised of the SoU and SoN. The annotations reveal how she touches on the collective values and identities of the communities she addresses through the SoU and paints a picture of a desired future (SoN and outcome) to call for healing and unity. In contrast, Trump's speech oscillates between short blocks of SoS and SoN with occasional SoU interspersed. In doing so, we can see how he leverages his personal experiences and the present-day challenges to rally his audience in a campaign context. Both speeches show the general flow of Challenge → Choice → Outcome, highlighting how the core PN elements can be adapted to different audiences, contexts, and goals while still reflecting a similar underlying narrative structure. Appendix~\ref{appendix:qual-analysis-political} contains a more thorough qualitative discussion and description of each.

This demonstrates the potential of our framework for application to related texts that possess characteristics of PNs, even if they were not crafted to conform strictly to their structural components, potentially enabling the further analysis of PN features more broadly.

\section{Discussion}

\subsection{Performance Varies Across Codes}
Our analysis of LLM annotation performance across codes reveals notable differences (Figure~\ref{f1boxplot}). On the more frequent codes, the model performed extremely well (SoS, SoU, SoN, challenge, call to action). These codes in particular may have been more clear to the model due to certain keywords that tend to signal these components (e.g. "I" for SoS, "together" or "we" for SoU, "now" for SoN and call to action, etc). On the other hand, the model struggles on more subjective codes: dream, nightmare, and urgency. While the results may be skewed due to low frequency in the dataset, these codes are often implied by the speaker, making them significantly less explicit compared to the categorical codes.
This may indicate that while LLMs are effective at analyzing individual story arcs, models struggle more with the rhetorical dimensions of narration, a key element of both civic, collective, and public narratives. This aligns with prior work \citep{chhun_language_2024} finding that when used for text analysis, LLMs favor surface-level (rather than inferential or relational) features. 

Overall, this highlights a fundamental challenge for LLMs in narrative annotation: while they can recognize explicit structural elements, they struggle with subjective or interpretative aspects of storytelling, which are paramount, by design, to the effectiveness of Public Narratives.

\subsection{Results Comparison to Related Works}
Human annotation is notoriously expensive and time-consuming \cite{carrell_is_2016}, making the use LLM-assisted annotation both reasonable and necessary. As we discovered in our codebook development process, wherein expert annotators frequently disagreed with one another's interpretations of elements of the established PN framework, PN annotation is decidedly more interpretative than "general content" coding. Nonetheless, we found that our models operated within expected ranges, with F1 scores aligning with prior work in related narrative, civic, and similarly subjective content annotations (e.g., \citealp{chew_llm-assisted_2023}); our use of \texttt{gpt-4o-mini} performing comparably to \citet{lupo_towards_2024}; and our use of \texttt{o3-mini} slightly outperforming past model applications.

We further relate our findings to the human-AI annotation study from \citet{lupo_towards_2024} using GPT-4 with similarly subjective content, a three-annotator evaluation pipeline, and comparable human-annotator and inter-run kappas. While our annotation task is more interpretative (evaluating mobilization potential vs. categorizing social roles), and though our models are slightly more advanced, \textit{our results corroborate the evidence of LLMs' utility for annotation tasks either alongside or in substitution for human annotators}. As with the prior study, we likewise found that precision generally outperformed recall, with LLMs better at identifying true coding instances than they were at accounting for missing ones. 

For any future work leveraging LLMs as textual annotators across qualitative domains, we present a summary of our takeaways that might be useful: 
\begin{itemize}
    \item LLMs are very capable at adhering to codebooks when given the exact detailed coding definitions (corroborates \citep{lupo_towards_2024,xiao-qual-2023}), and this performance will only increase as models continue to improve.
    \item While SOTA LLMs boast large context windows, annotation accuracy increased when we split the task into three chained prompts as opposed to one prompt to annotate a larger number of codes. We suggest experimenting with decreasing the complexity of the annotation whenever possible to increase performance.
    \item Few-shot examples did not improve performance (corroborates \citep{xiao-qual-2023}), causing models to overly adhere to the examples instead of codebook definitions.
    \item The models sometimes relied on keywords based on the code names instead of strictly adhering to the code definitions provided in the codebook. This was alleviated by giving more context in the system prompt regarding how to interpret the codebook in our domain.
\end{itemize}

\section{Conclusion} 

In this work, we iteratively develop a codebook to systematically annotate PNs. We create a dataset of 22 PNs and collect annotations from two domain experts on eight of them.  Then, we test to what extent highly capable LLMs are able to perform the annotation task and find high agreement with the human annotations. We perform an exploratory empirical analysis of PNs to investigate the extent to which real narratives align with PN theory. Moreover, we extend our methods to two recent political speeches, demonstrating the wider applicability of our codebook and LLM implementation to conduct civic narrative analyses. Our novel framework and empirical findings lay the groundwork for future large-scale evaluations of PNs to deepen our understanding of how leaders leverage the unique features of the PN framework how that translates to real-world impact in civic spaces.

Moreover, results from our study support  past work using LLMs for text annotation tasks, reinforcing both their potential and limitations for augmenting or replacing human annotators. Specifically, our work corroborates past studies (e.g., \citealp{ahmed_can_2024,lupo_towards_2024}) to demonstrate that although LLMs can achieve human-level performance, they struggle on tasks requiring contextual inference and nuanced understanding.
Our work contributes to an ongoing and necessary dialogue on best practices for LLM-assisted annotations, emphasizing the importance of both structured codebook frameworks and task-specific validation procedures.

\section{Limitations \& Future Work}

Through this work, we identify several areas for future work as well as possible limitations in the scope of our analysis. Due to resource constraints, we were not able to test a wider range of LLMs of more diverse sizes, developed by different companies, or open source models. Moreover, there are infinitely more prompting techniques and other improvements that could be experimented with to improve performance on specific codes. For example, it is likely that tweaking the codebook definitions specifically for LLM prompts could improve performance, which is unfortunately out of the scope of this work. As such, we do not claim that we have achieved the best possible performance, but rather see our work as discovering an acceptable lower bound of LLM performance with minimal prompt engineering. 

Second, beyond F1 scores, there may be additional metrics for comparison of LLM annotations with human experts that better account for the nuance and diversity in human interpretation of the narratives. For example, future work could explore manual validation of LLM annotations wherein the expert could use their judgement on whether the instances of disagreement are validly subjective or objective violations of the codebook. To scale this, it could even be possible to peform this using the LLM-as-a-Judge paradigm \citep{zheng_judging_2023}.

Regarding the political speeches, while we demonstrate that our framework lends an interesting new analysis perspective, it is important to note that political speeches differ significantly from the coached PNs we collected in this work. While our annotations indicate that a large portion of the Trump and Ardern speeches can be categorized via the PN codebook, we acknowledge that the instances of each element in the political speeches may be "less pure" examples of these elements compared to the true PNs. 
This discrepancy points to a limitation of our work: there is currently no differentiation between the quality or strength of the codes in our binary coding scheme. We hope that future work can extend the codebook to a more fine-grained ordinal scale to capture these nuances and open up further avenues of analysis.

\bibliography{custom}

\newpage
\appendix
\onecolumn
\section{Public Narrative Details}\label{appendix:pn}
A well-told story enables past moments to be experienced in the present, or a distant moment to be experienced as proximate, through a process known as narrative transportation \cite{appel_transportation_2010}. The more specific, sensory, and visual, the more the story might feel real and be emotionally accessible. Hope, for example, is central to effective storytelling. Experiences of loss or hurt often serve as motivation for an individual's care for a given cause or development of personal values. Likewise, hopefulness enables storyteller and listener to embrace "possibility" rather than be constrained by "probability" (a space between certainty and fantasy). 
Each of these components forms the basis of our coding scheme, which we designed to analyze PNs systematically. While there is much more to the PN framework—the construction of "narrative moments" serving as fundamental units of narration—in the scope of the present work, we focus principally on the framework's high-level components. 

\section{Qualitative Analysis of Political Speeches}\label{appendix:qual-analysis-political}
We analyzed two public speeches to identify components of Public Narrative. In addition to an automated annotation, we performed manual qualitative analysis on each speech. 

\textbf{\textit{Jacinda Ardern, New Zealand PM:}} Ardern's speech (March, 2019) is best categorized as a collective resilience narrative that builds a collective "Us" by way of empathetic reflection and identity-making. She begins and ends her speech in the Indigenous Māori language of Aotearoa New Zealand, and scaffolds her speech around mentions of "As-salaam Alaikum," or, "Peace be upon you" in Arabic (the language of those killed, and the greeting given in the days that have followed). This greeting echoed the values that Ardern asks people to retain as they move forward: love, peace, family. Ardern's speech is primarily centered around building cohesion across smaller communities to paint a picture of a diverse and collective "Us," New Zealanders, who are all affected by the events, and must all take up the mantle of hope. Ardern repeatedly references the specific and diverse stories of those who were impacted, overtly referencing the collective and the responsibility that comes along with it. She calls for people to live their shared values, and uses language that emphasizes shared identity.

\textbf{\textit{Donald Trump, Presidential Candidate:}} Trump's speech (July, 2024) focuses on his heroism, his experience, and his Story of Self. He uses imagery to place listener's in the moment of the story. He emphasizes his strength, the support he has received from Americans. He repeatedly emphasizes his past successes on immigration and economics. As he shares his story, 12 large pictures of his speech are projected on 12 monitors that surround him at the event. A backdrop of the White House is behind him. Trump reflects that God was on his side: he took an action (moving his head in a particular way) that he repeatedly tied to God's will to save his life. Additional pictures are shown of the blood dripping from his face. He compliments the crowd that was present, saying they were brave for not running when they heard the gunshots, which he said is typical of mass-shootings (people running). He calls his crowd smart, saying that they knew immediately what kind of gun it was (e.g., sniper), and were immediately looking for the shooter, bravely, instead of trying to save themselves. He closes his speech, again, by calling it God's providence that he survived. He recalls the moment when he calls for his people to fight. Notably, the images of blood dripping off Trump's face remained up for the entirety of the speech, except when he closes with God's providence line, and shows a picture of him holding a fist up, embodying strength and calling for his followers to "fight," a word that they then chant.

\textbf{\textit{Cross-Comparison:}} Both speeches embody key elements of Public Narrative, but in very different ways that we believe reflect differences in the function of each speech. Trump's speech is dominated by Stories of Self and Now, connecting his past actions to the future sought by listeners—positioning his narrative as functional towards his then-candidacy. Ardern's speech is instead driven largely by Story of Us and Now, connecting the actions of others, and by a larger collective population, to hopeful outcomes in the future—functioning to leverage collectivism to motivate healing. While Ardern calls on the collective to recover, and heal, Trump calls instead on his collective to fight. Trump celebrates the death of the shooter at his rally, whereas Ardern celebrates the values of survivors. Values appear in both speeches, but are more religious in Trump's speech, and more family- and collective-based in Ardern's (e.g., love). Both sets of values, we note, are pro-social, but the pro-social values of the former's speech appear only inclusive of particular groups (e.g., Christians), whereas Ardern's are inclusive of a larger array of individual identity groups. Trump places himself at the center of his community, whereas Ardern diffuses the community to comprise a wide array of individuals, never mentioning herself explicitly. There is no real Story of Self (SoS) in Ardern's speech, but Trump's speech is almost entirely SoS, with Trump focusing almost exclusively on the past and his own deeds and heroism. The crowd during Ardern's speech was exceptionally quiet compared to those of Trump's, who cheered and chanted (e.g., "Fight!") repeatedly. While imagery (e.g., blood, American flags, White House backdrop, etc.) was used during Trump's speech, no imagery was clearly used during Ardern's.

\section{LLM-Annotated PN Exploratory Analysis}\label{analysis}
\begin{table}[h]
\centering
\begin{tabular}{lc}
\toprule
Code            & Overall Average (\%) \\
\midrule
self            & 48.61 \\
story details    & 47.37 \\
now             & 30.77 \\
challenge       & 27.02 \\
us              & 24.34 \\
outcome         & 19.65 \\
choice          & 16.83 \\
vulnerability   & 16.23 \\
values          & 14.57 \\
call-to-action    & 11.21 \\
hope            & 10.63 \\
urgency         &  4.23 \\
nightmare       &  3.76 \\
dream           &  1.43 \\
\bottomrule
\end{tabular}
\caption{Overall average percentage of each narrative code across the 22 LLM-annotated workshop speeches.}
\label{appendix:desc-stats}
\end{table}

\begin{figure}[ht]
    \centering
    \includegraphics[width=0.7\linewidth]{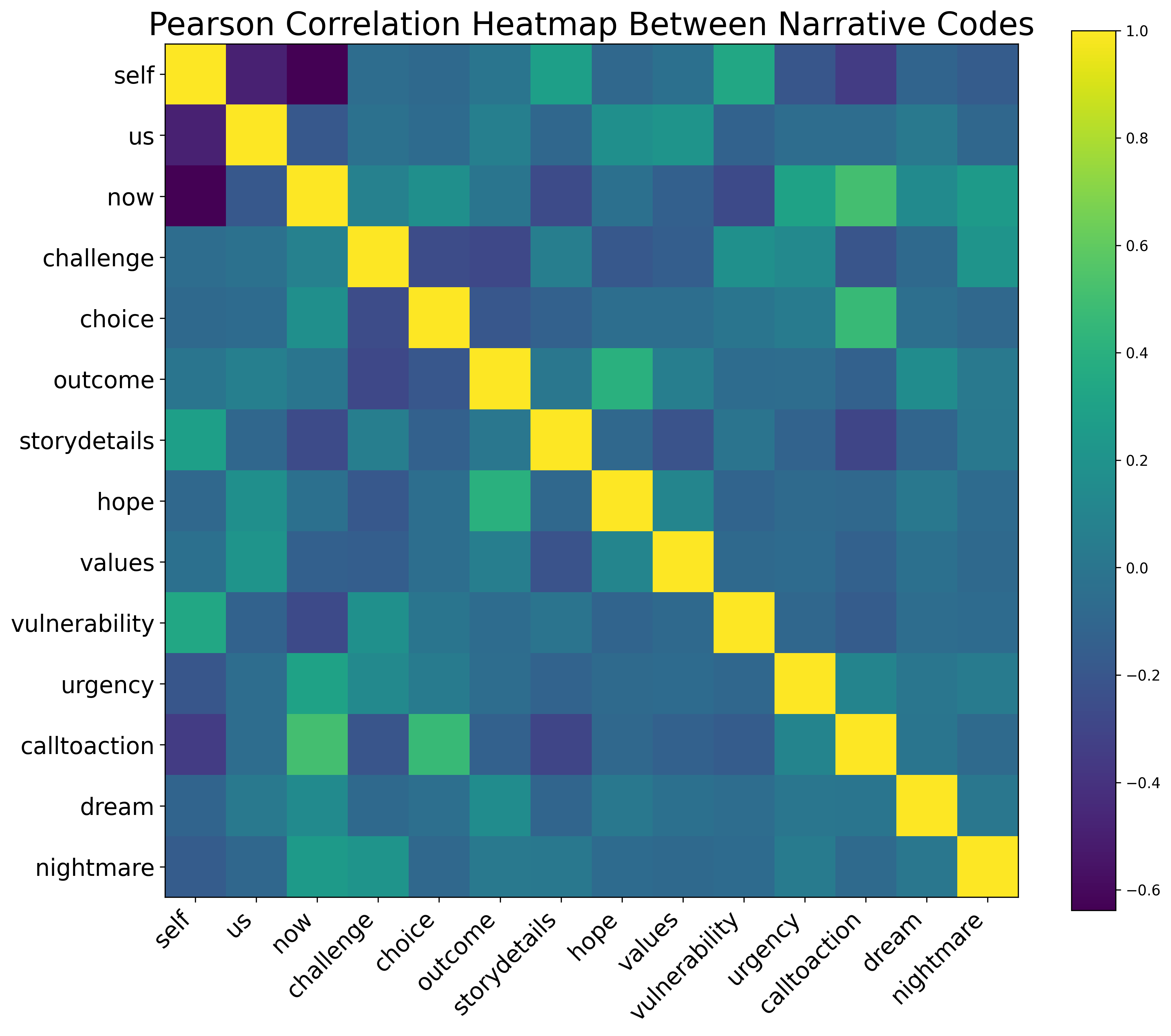}
    \caption{Pearson correlation heatmap of code co-occurrence for \texttt{o3-mini} with \textsc{CoT + Prompt Chaining}, averaged across the 22 PNs.}
    \label{pearson-heatmap}
\end{figure}

\begin{figure}[ht]
    \centering
    \includegraphics[width=0.7\linewidth]{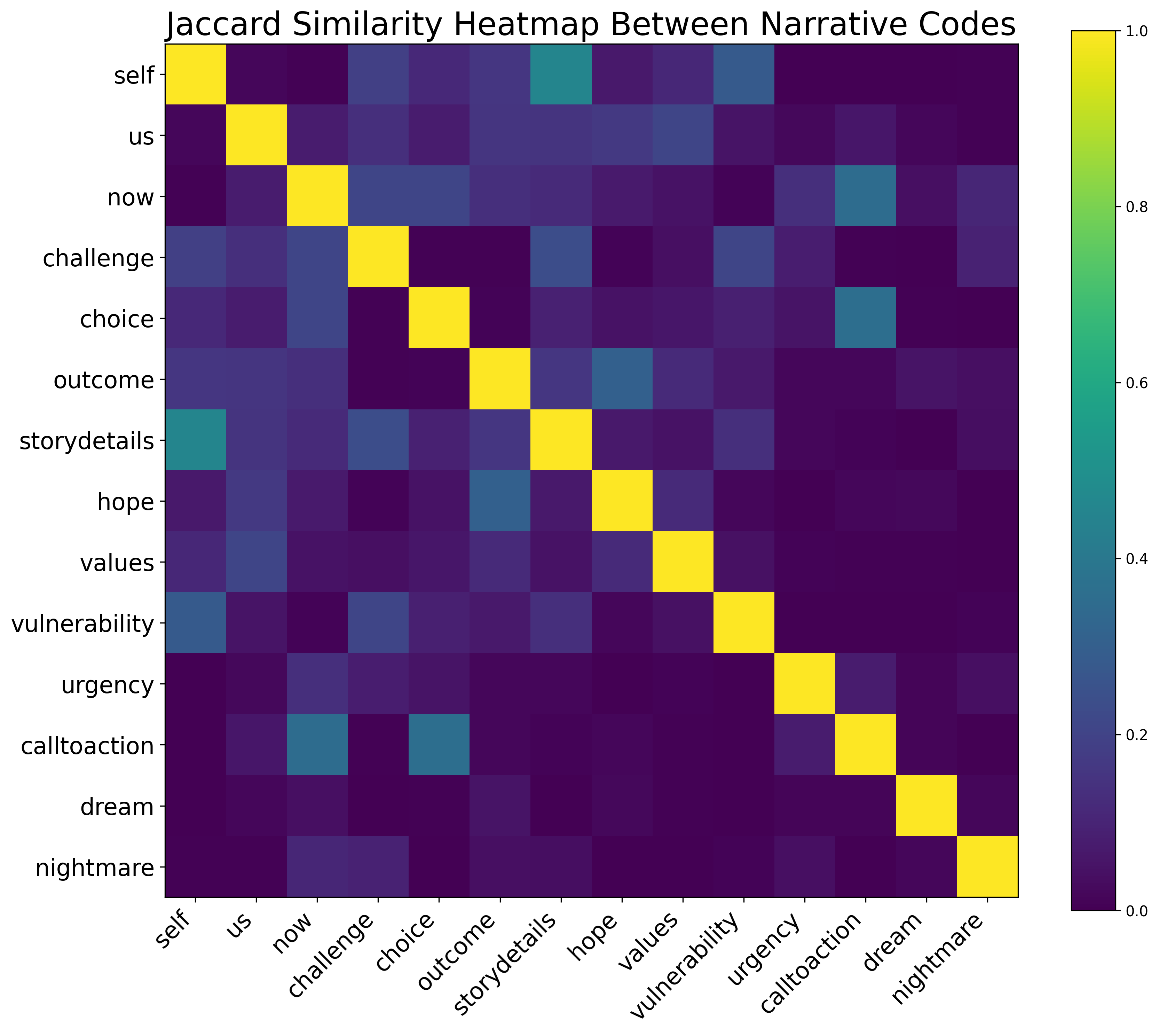}
    \caption{Jaccard similarity heatmap of code co-occurrence for \texttt{o3-mini} with \textsc{CoT + Prompt Chaining}, averaged across the 22 PNs.}
    \label{jaccard-heatmap}
\end{figure}

\subsection{Discussion of Code Frequency}\label{freq}
Across the 22 LLM-annotated workshop narratives, certain codes appeared with greater frequency than others. The most prevalent code was SoS (48.6\%), reflecting the importance of personal storytelling in participants' public narratives. Story details were also highly present (47.4\%), which suggested that participants frequently used specific context, vivid sensory details, or characterizations to ground their narratives. These features were followed by SoN (30.77\%), challenge (27.02\%), and SoU (24.34\%), reflecting the usage of temporal urgency, tension, and collective framing in shaping PNs. However, emotional or motivational appeals such as vulnerability (16.2\%), values (14.6\%), and hope (10.6\%) were much less dominant. Content-related and more affective codes like urgency (4.2\%), nightmare (3.8\%), and especially dream (1.4\%) were annotated sparingly, suggesting these abstract or aspirational elements were less consistently invoked across the workshop narratives. The disparity between structural and categorical vs. content codes supports the argument that just as structural elements may be more consistently legible to human and LLM annotators, affective features like hope and urgency may surface more subjectively and thus pose greater difficulty for consistent identification.

\subsection{Code Co‐occurrence Similarity}\label{jaccard}
To examine how codes co‐occur within the same sentences, we computed pairwise Jaccard similarities across all 14 codes for the 22 transcripts (see Figure~\ref{jaccard-heatmap}). This metric reflects the proportion of lines in which two codes appear together out of all lines in which either appears:

$$J(A,B) = \frac{A\cap B}{A\cup B}$$

The strongest co‐occurrence was between SoS and storydetails ($J = 0.455$), suggesting that personal storytelling nearly always includes rich narrative detail—an observation consistent with their high individual frequencies noted earlier. This finding reinforces the idea that narrators ground individual experience through contextual specifics, and mirrors the prominence of both codes discussed in the frequency analysis.

Several high-similarity pairings also echoed the strongest Pearson correlations. For example, choice and call to action ($J = 0.361$) and SoN and call to action ($J = 0.354$) were among the top Jaccard pairs, aligning closely with their respective Pearson values ($r = 0.468$ and $r = 0.513$). This cross-metric agreement strengthens the conclusion that temporal framing (SoN) and decision-making (choice) frequently precede or accompany explicit prompts for action in public narratives. Outcome and hope ($J = 0.305$) also co-occurred meaningfully, reinforcing their observed moderate correlation ($r = 0.404$) and suggesting that expressions of results often carry a forward-looking emotional tone.

At a moderate level ($0.2 \leq J < 0.3$), we observe additional thematic pairings that provide further nuance:

\begin{itemize}
\item self \& vulnerability ($J = 0.282$): personal-experience lines often reveal emotional openness.
\item challenge \& story details ($J = 0.236$): accounts of conflict are frequently embedded in narrative specifics.
\item now \& challenge ($J = 0.210$): temporal immediacy is often used to present obstacles.
\item now \& choice ($J = 0.209$): time-bound stakes frequently set up decision points.
\item challenge \& vulnerability ($J = 0.208$): narrators often frame challenges as emotionally resonant experiences.
\item us \& values ($J = 0.207$): collective identity appeals often evoke shared moral frameworks.
\end{itemize}

Interestingly, while the Pearson analysis revealed strong negative associations—such as between SoS and SoN ($r = -0.638$), and between SoS and SoU ($r = -0.486$)—these inverse relationships are not captured by Jaccard similarity, which only considers joint presence. This contrast illustrates the difference between measuring association (via correlation) and co-occurrence (via Jaccard): the former captures whether codes tend to appear in opposition or together, while the latter strictly quantifies overlap when either is present.

Finally, no code pair exceeded a Jaccard index of 0.5, indicating that even the most commonly co-occurring codes are not overwhelmingly inseparable on a line-by-line basis. This underscores the flexibility and combinatory nature of code application across different narrative contexts.

\section{Human Annotator Agreement}
\begin{table}[ht]
\centering
\begin{tabular}{lcccc}
\toprule
Code & Avg. Freq & $p_o$ & PABAK & $\kappa$ \\
\midrule
Story of Self & 222.0 & 0.86 & 0.73 & 0.73 \\
Story of Us & 70.0 & 0.88 & 0.75 & 0.53 \\
Story of Now & 187.5 & 0.86 & 0.72 & 0.71 \\
Challenge & 141.5 & 0.80 & 0.60 & 0.53 \\
Choice & 85.5 & 0.80 & 0.60 & 0.35 \\
Outcome & 80.0 & 0.77 & 0.54 & 0.22 \\
Story Details & 51.5 & 0.88 & 0.77 & 0.42 \\
Hope & 48.0 & 0.84 & 0.67 & 0.16 \\
Values & 106.0 & 0.68 & 0.36 & 0.19 \\
Vulnerability & 78.0 & 0.80 & 0.59 & 0.30 \\
Urgency & 72.5 & 0.77 & 0.54 & 0.19 \\
Call-to-Action & 67.5 & 0.89 & 0.78 & 0.58 \\
Dream & 15.0 & 0.95 & 0.89 & 0.17 \\
Nightmare & 5.5 & 0.98 & 0.96 & 0.17 \\
\midrule
Global (micro) & --- & 0.84 & 0.68 & 0.49 \\
Macro-average & --- & --- & 0.68 & 0.38 \\
\bottomrule
\end{tabular}
\caption{Inter-annotator agreement for 14 binary codes across eight stories (two annotators each).  
Avg.\ Freq.\ is the average number of positive annotations per code; $p_o$ is the raw percent agreement; PABAK ($2p_o\!-\!1$) is the prevalence- and bias-adjusted kappa; and $\kappa$ is Cohen's kappa.  
"Global (micro)" reports each statistic computed on the flattened set of all code–line decisions; "Macro-average" is the mean of the 14 per-code values. A dash (---) indicates not applicable.}\label{tab:agreement}
\end{table}

\subsection{Discussion on Subjectivity and IIA}\label{appendix:annotator-agreement}
Notably, the relationship between categorical elements of the second exemplar narrative were more traditionally established (Self → Us → Now), which may have supported more consistent human interpretations across codes. This was true even though this story's structural codes (Choice → Challenge → Outcome) appeared more sporadically distributed. As described by our third annotator, and verified in our analysis, Choice codes appeared sporadically throughout this story, rather than appearing as a structural "stage" or singular "moment" within it. Nonetheless, the high-level structure afforded by linear categorical elements of the narrative may have enhanced human agreement across codes and coding groups. These scores suggest an inherent subjectivity in the process of annotating PNs.

\section{Model and Prompt Selection}
\subsection{Prompt Descriptions}\label{appendix:prompting}
Below are the three prompt techniques we tested:
\begin{enumerate}
    \item \textsc{Chain of Thought} (\textsc{CoT}; \citealt{wei_chain_2022}): The prompt contains the definitions of the 6 codes with no examples, the LLM must output annotations for 6 codes using CoT.
    \item \textsc{CoT + Few-Shot} \citep{brown_fewshot_2020}: The prompt contains the definitions of the 6 codes plus a coding example from the codebook for each code. The LLM must output annotations for 6 codes using CoT.
    \item \textsc{CoT + Prompt Chaining} \citep{wu-chaining-2022}: Prompt chaining involves dividing a complex task into several smaller tasks, where the LLM output of a previous task becomes an input for the following prompt. In our case, we have three sub-tasks which more closely reflects the human annotation process, in which categorical codes are annotated prior to the rest:
    \begin{enumerate}
        \item The first prompt contains the definitions of the 3 categorical codes with no examples. The LLM must output annotations for just these codes using CoT.
        \item The second prompt contains the definitions of the 3 structural codes, no examples, and its annotations from the output of the first query. The LLM must output annotations for the 3 structural codes using CoT.
    \end{enumerate}
\end{enumerate}
\subsection{Results}\label{appendix:more-results}

This section contains the results on the two exemplar PNs used for teaching that were used in our codebook development. Due to their use as teaching examples, the two narratives are quite different and as such we analyzed them separately, denoting one as "A" and the other as "B." Their differences allowed us to more deeply understand the nuances in how speakers can effectively use the PN framework in diverse ways and ensure that our model and prompt selection could account for this. In reality, the PNs in our dataset (Section~\ref{datacollection}) are on the simpler side and reflect more of the structure of "B."

\textit{Note}: Unless otherwise stated, mentions of F1 scores in this section are averaged across both narratives. 

Overall, we see that \textsc{CoT + Prompt Chaining} achieves the best performance for \texttt{gpt-4o-mini} predicted (mean F1 on A=$0.57$; B=$0.55$), whereas \textsc{CoT} (mean F1 on A=$0.49$; B=$0.55$) and \textsc{CoT + Few Shot} (mean F1 on A=$0.50$; B=$0.49$) are slightly lower. 
For \texttt{o3-mini}, all three prompts achieve almost identical performance \texttt{o3-mini} (mean F1 on A=$0.73\pm0.0$; B=$0.61\pm0.01$). Similarly for \texttt{gpt-4o-mini} structured, all three prompts perform the about same (mean F1 on A=$0.49\pm0.02$; B=$0.50\pm0.01$), but it is overall the worst performing model.

\begin{figure*}[t]
    \centering
    \includegraphics[width=\linewidth]{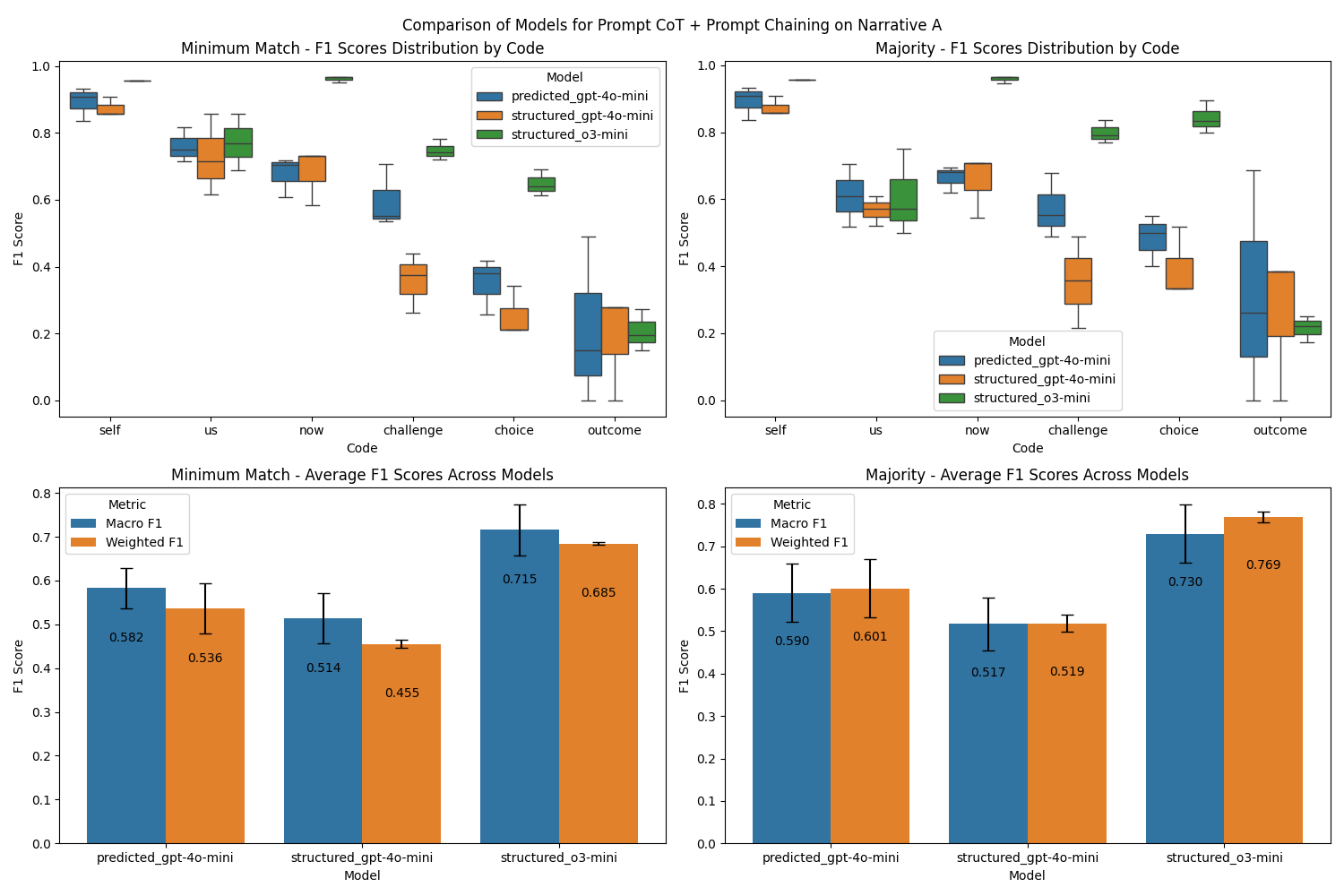}
    \caption{Performance of each model with \textsc{CoT + Prompt Chaining} averaged across 3 runs.
    }
    \label{best-prompt-all-models}
\end{figure*}

Figure~\ref{best-prompt-all-models} shows the performance of each model with \textsc{CoT + Prompt Chaining} for narrative A along with the code-level F1 scores. Here, we see that \texttt{o3-mini} is clearly the best model overall, achieving on average 0.20 higher F1 scores than \texttt{gpt-4o-mini} structured and 0.11 higher than \texttt{gpt-4o-mini} predicted. 

\begin{figure*}
    \centering
    \includegraphics[width=1\linewidth]{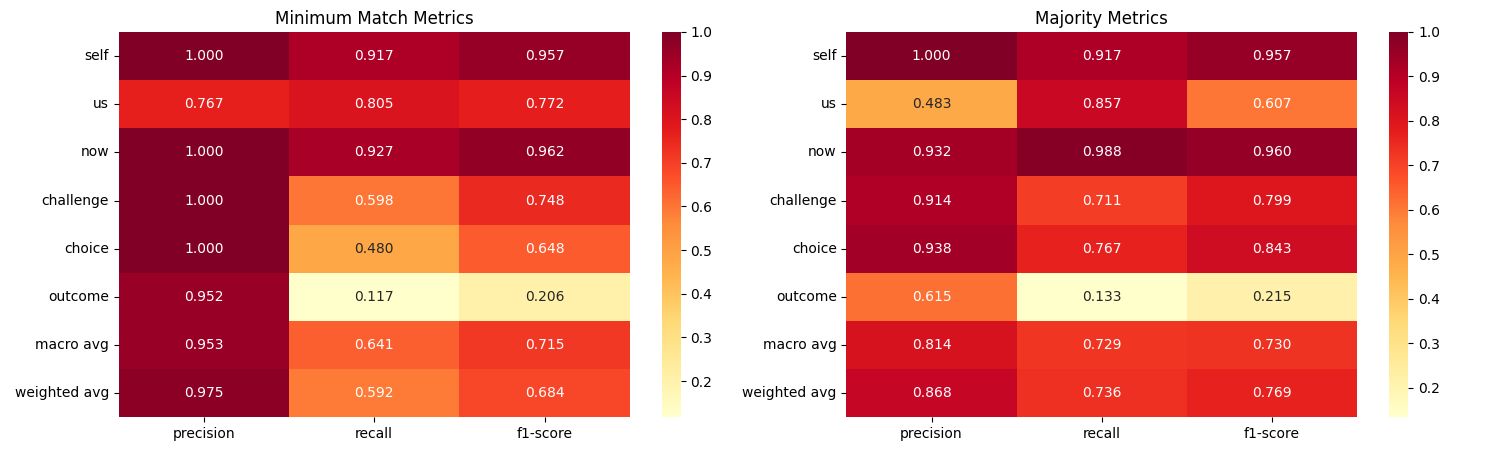}
    \caption{Detailed results for \texttt{o3-mini} with \textsc{CoT + Prompt Chaining} averaged across three runs.  
    }
    \label{o3-james-v4}
\end{figure*}

Figure~\ref{o3-james-v4} shows the detailed breakdown of \texttt{o3-mini} performance with \textsc{CoT + Prompt Chaining} on narrative A. Across both narratives A and B, we observe that between the three categorical codes, \texttt{o3-mini}'s performance is extremely high on SoS and SoN (both mean F1=$0.96$) and lower on SoU (mean F1=$0.69\pm0.08$). Among the three structural codes, the model does well on Challenge (mean F1=$0.77\pm0.02$) and Choice (mean F1=$0.75\pm0.10$) and struggles with Outcome (mean F1=$0.21\pm0.01$). These trends are also observed for both \texttt{gpt-4o-mini} structured and predicted across all prompts and both narratives.

\begin{figure*}[h]
    \centering
    \includegraphics[width=\linewidth]{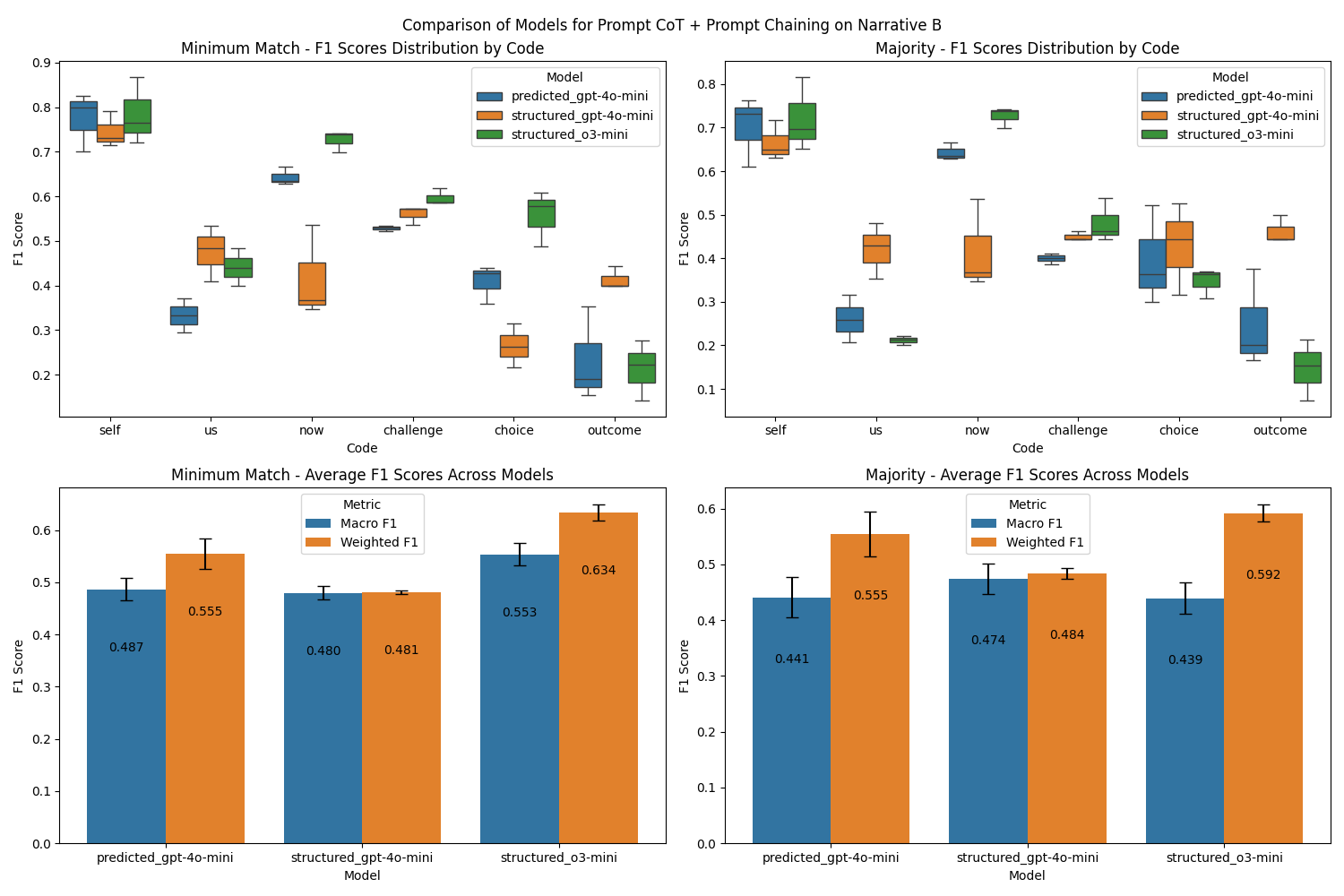}
    \caption{Performance of each model using the CoT + Prompt Chaining prompt averaged across 3 runs.}
    \label{maung_v4}
\end{figure*}

\begin{figure*}[h]
    \centering
    \includegraphics[width=\linewidth]{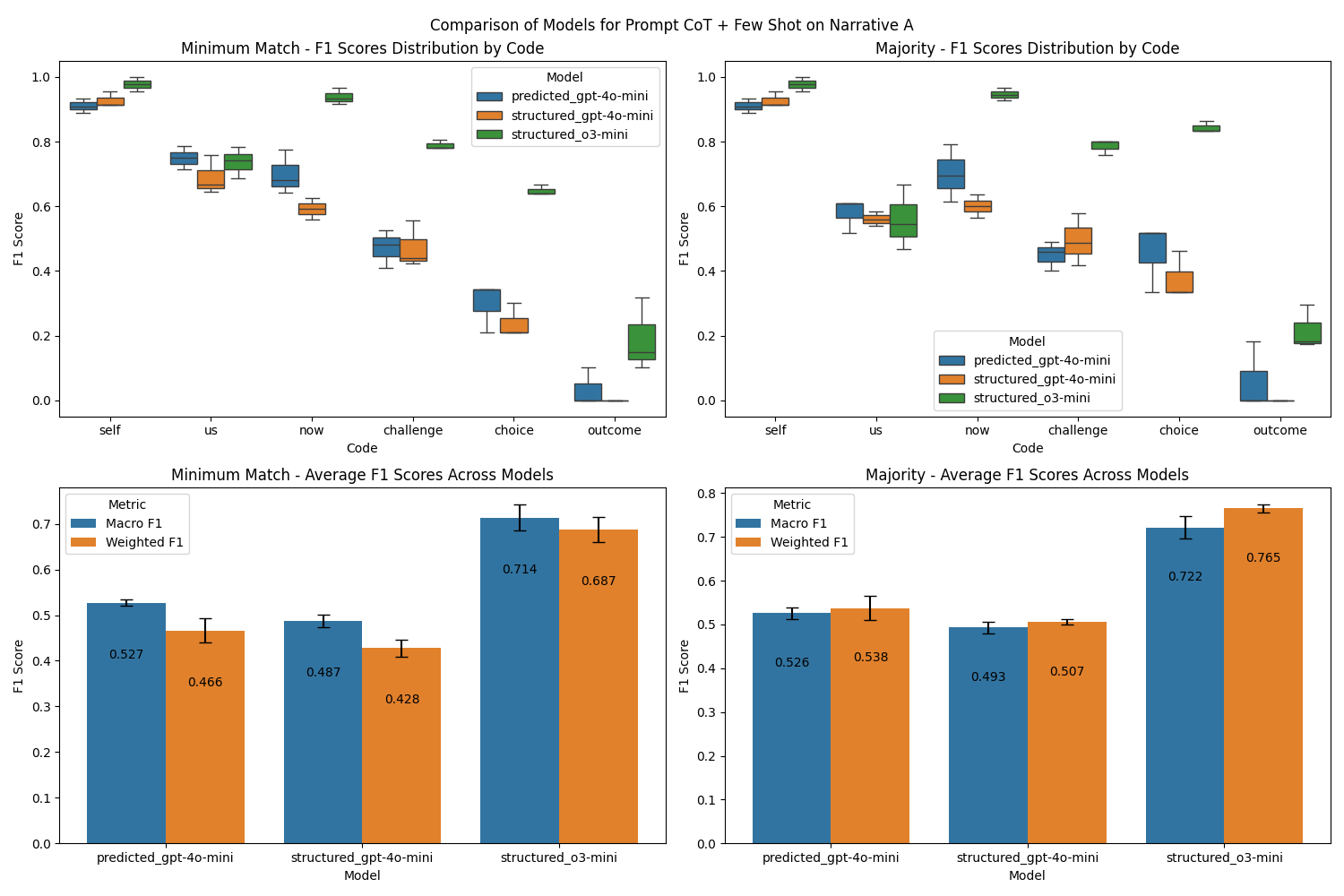}
    \caption{Performance of each model using the CoT + Few Shot prompt averaged across 3 runs Narrative A.}
    \label{james_v3}
\end{figure*}

\begin{figure*}[h]
    \centering
    \includegraphics[width=\linewidth]{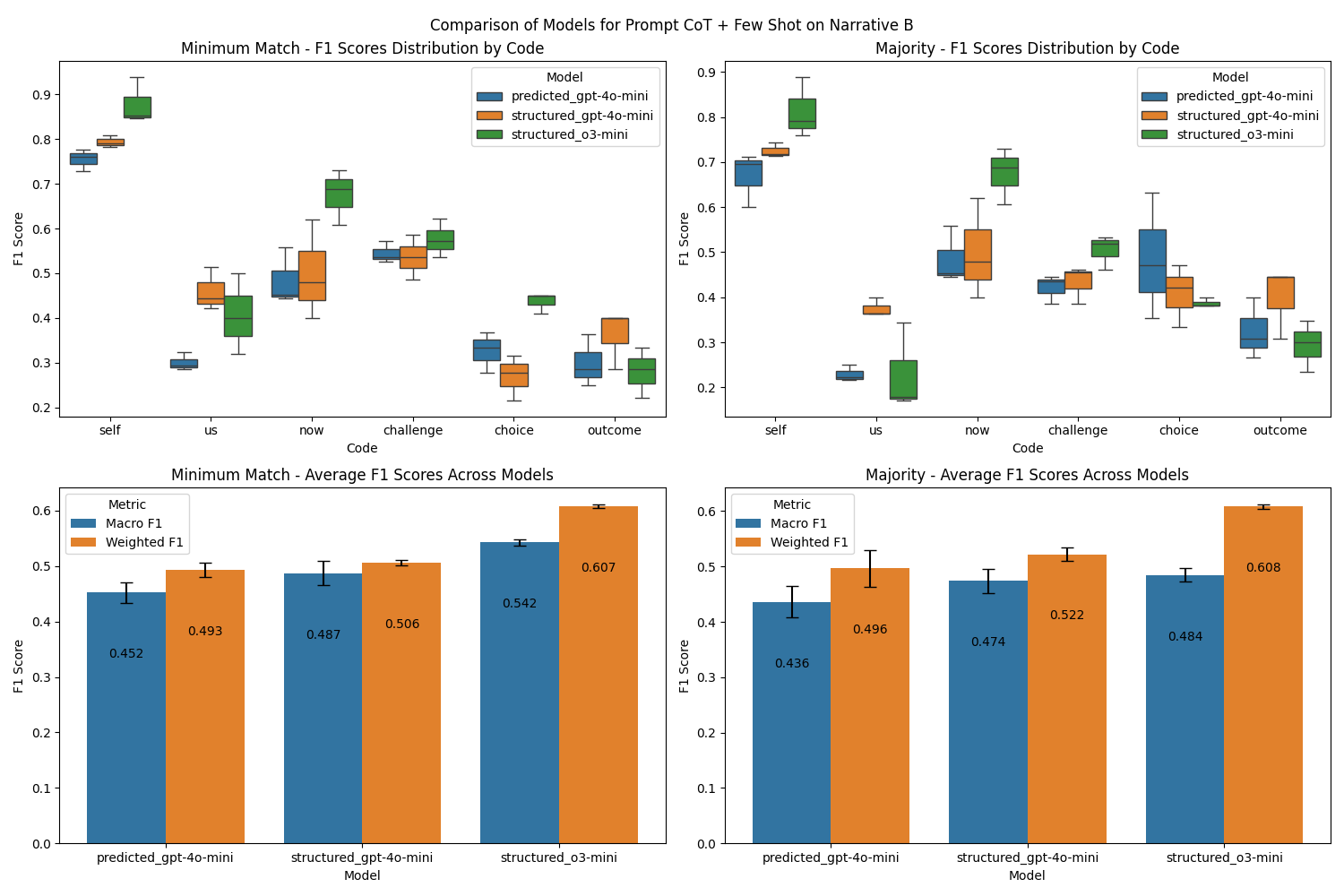}
    \caption{Performance of each model using the CoT + Few Shot prompt averaged across 3 runs Narrative B.}
    \label{maung_v3}
\end{figure*}

\begin{figure*}[h]
    \centering
    \includegraphics[width=\linewidth]{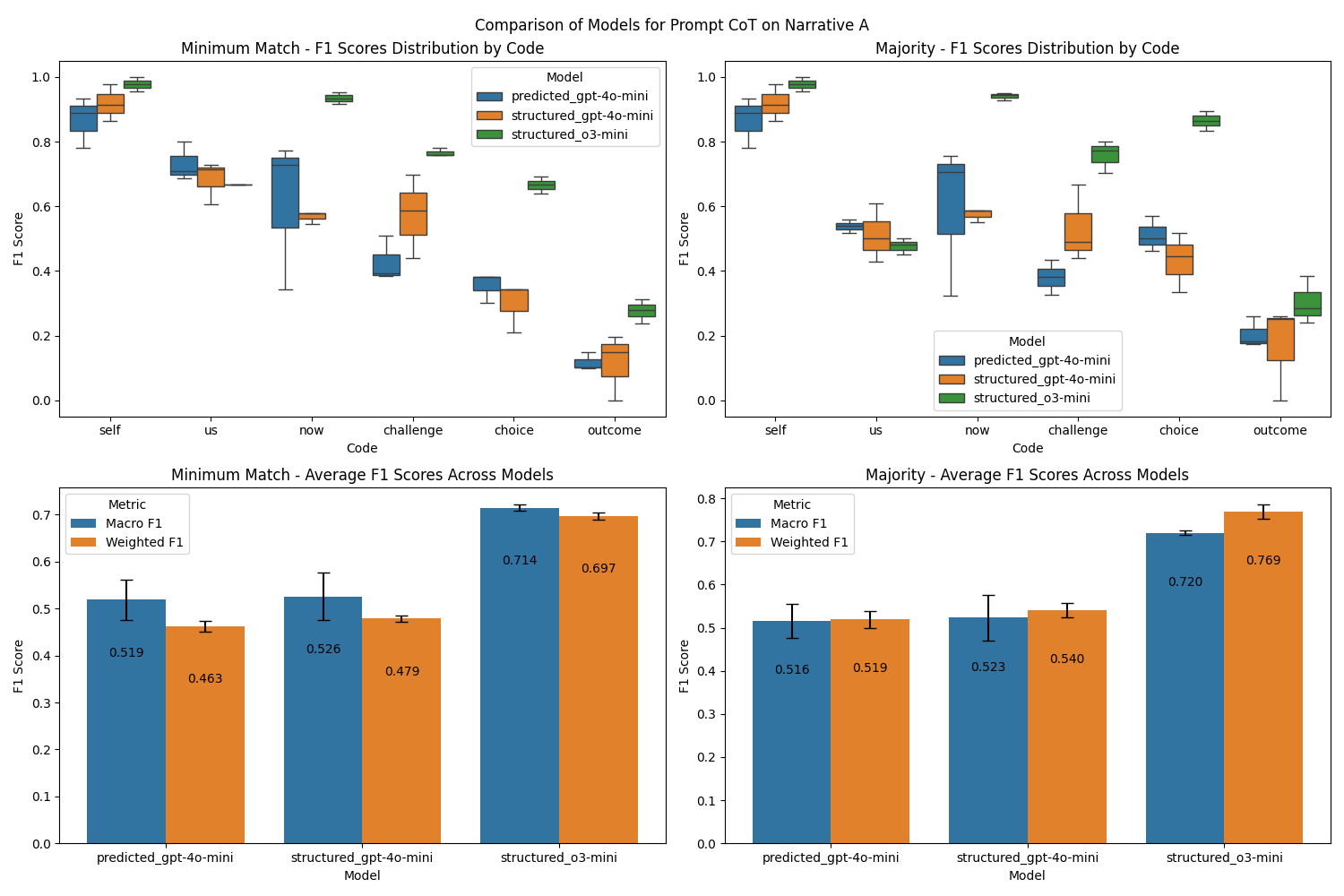}
    \caption{Performance of each model using the CoT  prompt averaged across 3 runs Narrative A.}
    \label{james_v2}
\end{figure*}

\begin{figure*}[h]
    \centering
    \includegraphics[width=\linewidth]{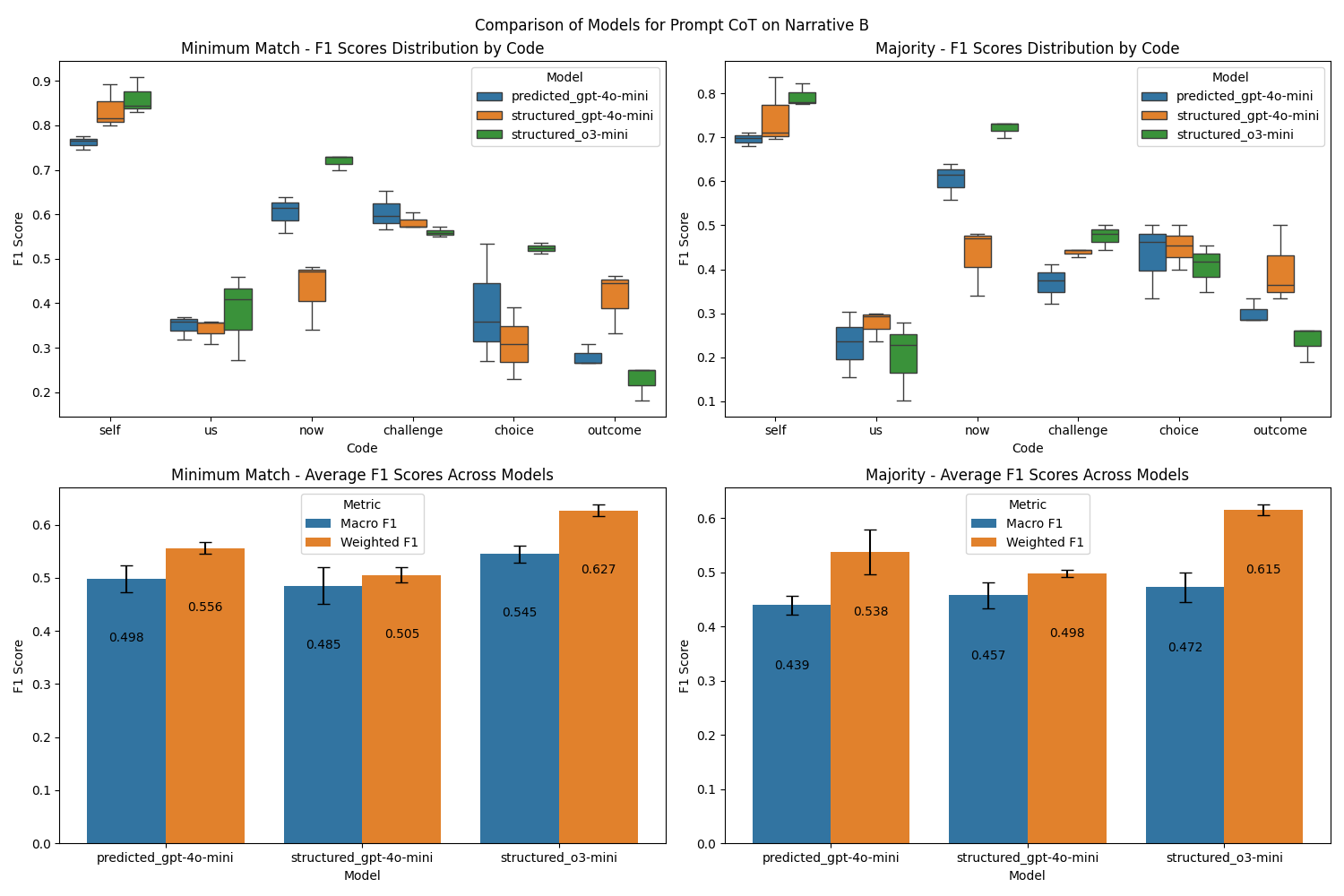}
    \caption{Performance of each model using the CoT prompt averaged across 3 runs Narrative B.}
    \label{maung_v2}
\end{figure*}

Figure~\ref{maung_v4} shows the analogous results to Figure~\ref{best-prompt-all-models} except on Narrative B. Figures~\ref{james_v3},~\ref{maung_v3},~\ref{james_v2}, and~\ref{maung_v2} show the performance of each model for \textsc{CoT + Few Shot} and \textsc{CoT} prompt experiments for narratives A and B, respectively.

\subsection{Minimum Match vs. Majority} 
We compare two evaluation methods for LLM annotation performance: Minimum Match, which considers a code valid if at least one human annotator assigned it, and Majority Match, which requires agreement from at least two out of three annotators.

Our results show that precision is generally higher and recall lower for Minimum Match (Figure~\ref{o3-james-v4}), as it allows more positive classifications, reducing false positives but increasing false negatives. Conversely, LLM performance compared to the Majority yields higher recall but lower precision, as fewer annotations meet the stricter agreement threshold. This pattern aligns with expectations: LLMs tend to agree with consensus labels while occasionally capturing subjective interpretations that only one annotator identified. 

Notably, this suggests that LLMs align well with human majority decisions but can also capture diverse narrative interpretations, an important consideration for subjective tasks like PN annotation. We view this as preferential, since narrative coding tasks are inherently subjective and interpretable and a diversity of perspectives in such cases should be actively sought rather than normalized through strict majority-capping.



\section{Prompting Experiment Prompts}
\label{appendix:prompts}

\subsection{CoT}
\subsubsection{CoT System Prompt}
\begin{verbatim}
Your task is to annotate a public narrative speech according to a specific codebook developed 
by Dr. Marshall Ganz from Harvard. Simply put, Public Narrative says, "Here's who I am, this 
is what we have in common, and here's what we're going to do about it." By mastering the 
practise of crafting a narrative that bridges the self, us, and now, organizers enhance their 
own efficacy and create trust and solidarity with their constituency.

The Public Narrative framework is made up of three components: a Story of Self, a Story of Us, 
and a Story of Now. A Story of Self communicates the values that have called you to leadership; 
a Story of Us communicates the values shared by those in action; a Story of Now communicates an 
urgent challenge to those values that demand action now. Each story within this framework 
follows a fundamental structure that grounds abstract values in concrete experiences: 
Challenge, Choice, and Outcome. Challenge shows a moment of adversity or uncertainty. Choice 
is the decision to respond to the challenge, shaped by the storyteller's values. An Outcome 
demonstrates the result of that choice, revealing the stakes and consequences of action (or 
inaction). For a Story of Self, this structure often unfolds through formative moments from 
one's youth or early leadership experiences. In a Story of Us, it highlights collective 
challenges, an "us's" shared decision, and their collective outcome. In a Story of Now, it 
underscores the present crisis and the imperative for action.

**Your Task**
Together, these six components (story of self, us, and now; challenge, choice, and outcome) 
form the structure of the narrative. Your task is to annotate a narrative and identify which 
parts belong to the story of self, us, and now, and which parts denote a challenge, choice, or 
outcome. Note that the narrative will not necessarily follow this order, and may switch between 
these components fluidly, so pay attention to the definitions below and always adhere to them 
when annotating:

**Narrative Code Definitions**
1. Story of Self: A personal experience that reveals the origin of one's values or why they are 
called to leadership. Sometimes called an Origin Story.
2. Story of Us: A communal experience that aims to highlight or generate a sense of shared 
values and experience within a group.
3. Story of Now: An experience that emphasizes the nature of current circumstances and aims 
to motivate an audience to take action to address it.

**Story Structure Code Definitions**
4. Challenge: A past, present, or potential future situation whose negative impact is to be 
overcome (or has been overcome). May affect either a collective or an individual.
5. Choice: A past, present, or possible future decision to be taken in response to a 
challenge, often reflecting personal values. May be individual or collective.
6. Outcome: A past, present, or possible future result of a choice, whether made or 
conditional. The consequences of this outcome could be known (positive or negative) or unknown, 
and they may affect either a collective or an individual (although more often a collective). 

You will output in a JSON format specified below.
\end{verbatim}
\subsubsection{CoT Prompt}
Below \texttt{\{narrative\}} is the full text of the input PN.
\begin{verbatim}
**Public Narrative:**
{narrative}

**Output Instructions**
Step 1. Consider the public narrative above and the definitions of the story of self, story of us, 
and story of now, and within this context, consider the definitions of challenge, choice, and 
outcome.
Step 2. Identify which sentences comprise the story of self, us, and now.
Step 3. Identify which sentences contain a choice, challenge, and/or outcome.
Step 4. Respond in JSON list for each sentence (don't skip any and do not edit the text), indicating 
a 1 in the respective field if the sentence is part of self/us/now and or contains a 
challenge/choice/outcome and 0 otherwise. Be sure to include ALL lines.
\end{verbatim}

\subsection{CoT + Few Shot}
\subsubsection{CoT + Few Shot System Prompt}
\begin{verbatim}
Your task is to annotate a public narrative speech according to a specific codebook developed 
by Dr. Marshall Ganz from Harvard. Simply put, Public Narrative says, "Here's who I am, this 
is what we have in common, and here's what we're going to do about it." By mastering the 
practise of crafting a narrative that bridges the self, us, and now, organizers enhance their 
own efficacy and create trust and solidarity with their constituency.

The Public Narrative framework is made up of three components: a Story of Self, a Story of Us, 
and a Story of Now. A Story of Self communicates the values that have called you to leadership; 
a Story of Us communicates the values shared by those in action; a Story of Now communicates an 
urgent challenge to those values that demand action now. Each story within this framework 
follows a fundamental structure that grounds abstract values in concrete experiences: 
Challenge, Choice, and Outcome. Challenge shows a moment of adversity or uncertainty. Choice 
is the decision to respond to the challenge, shaped by the storyteller's values. An Outcome 
demonstrates the result of that choice, revealing the stakes and consequences of action (or 
inaction). For a Story of Self, this structure often unfolds through formative moments from 
one's youth or early leadership experiences. In a Story of Us, it highlights collective 
challenges, an "us's" shared decision, and their collective outcome. In a Story of Now, it 
underscores the present crisis and the imperative for action.

**Your Task**
Together, these six components (story of self, us, and now; challenge, choice, and outcome) 
form the structure of the narrative. Your task is to annotate a narrative and identify which 
parts belong to the story of self, us, and now, and which parts denote a challenge, choice, or 
outcome. Note that the narrative will not necessarily follow this order, and may switch between 
these components fluidly, so pay attention to the definitions below and always adhere to them 
when annotating:

**Narrative Code Definitions + Examples **
1. Story of Self: A personal experience that reveals the origin of one's values or why they are 
called to leadership. Sometimes called an Origin Story.
    Gold standard example: A story moment with sensory details (e.g., "When I was 8 years old, my 
    grandmother took me to a protest for the first time. As we marched together, holding hands and 
    chanting, I felt a surge of power I'd never experienced before. That day, I learned that my
    voice mattered and we could make change.")
2. Story of Us: A communal experience that aims to highlight or generate a sense of shared values 
and experience within a group.
    Gold standard example: A story moment showing both shared values and collective action 
    (e.g., "Last year, when the factory threatened to close, we all felt scared and powerless. 
    But then Maria stood up at the union meeting and said, 'We've faced tough times before, 
    and we've always come through together.' Within a week, we had organized a campaign that 
    kept the factory open.")
3. Story of Now: An experience that emphasizes the nature of current circumstances and aims to 
motivate an audience to take action to address it.
    Gold standard example: "Right now, our children are struggling to read at grade level, with 
    40% of third graders falling behind. Current projections estimate that this challenge will 
    only exacerbate if left unaddressed. But there's hope. By volunteering just two hours a 
    week, each of us can help a child unlock the power of reading. Who will join me today?"

**Story Structure Code Definitions**
4. Challenge: A past, present, or potential future situation whose negative impact is to be 
overcome (or has been overcome). May affect either a collective or an individual.
    Gold standard example: Vivid depiction of a challenge with emotional impact (e.g., "The 
    doctor's words hit me like a ton of bricks: 'Your son has autism.' I felt completely lost 
    and overwhelmed, unsure of how to help my child navigate a world that suddenly seemed so 
    much more complicated.")
5. Choice: A past, present, or possible future decision to be taken in response to a challenge, 
often reflecting personal values. May be individual or collective.
    Gold standard example: Detailed account of a choice, including internal struggle and values 
    considered (e.g., "I could have walked away when I saw the bullies picking on the new kid. 
    It would have been easier. But I remembered how it felt to be alone and afraid, so I stepped 
    between them, saying, 'Hey, leave him alone. He's with me.'")
6. Outcome: A past, present, or possible future result of a choice, whether made or 
conditional. The consequences of this outcome could be known (positive or negative) or 
unknown, and they may affect either a collective or an individual (although more often a 
collective). 
    Gold standard example: Explanation of both immediate and long-term impacts, connecting to 
    personal growth or wider change (e.g., "After we decided to start the community gArdern, 
    not only did we have fresh vegetables, but I saw neighbors talking to each other for the 
    first time in years. We had created more than just a gArdern; we had grown a community.")

You will output in a JSON format specified below.
\end{verbatim}

\subsubsection{CoT + Few Shot Prompt}
The prompt for this is identical to that of the previous (CoT).

\subsection{CoT + Prompt Chaining}
\subsubsection{CoT + Prompt Chaining System Prompt 1}

\begin{verbatim}
Your task is to annotate a public narrative speech according to a specific codebook developed 
by Dr. Marshall Ganz from Harvard. Simply put, Public Narrative says, "Here's who I am, this 
is what we have in common, and here's what we're going to do about it." By mastering the 
practise of crafting a narrative that bridges the self, us, and now, organizers enhance their 
own efficacy and create trust and solidarity with their constituency.

The Public Narrative framework is made up of three components: a Story of Self, a Story of Us, 
and a Story of Now. A Story of Self communicates the values that have called you to leadership; 
a Story of Us communicates the values shared by those in action; a Story of Now communicates an 
urgent challenge to those values that demand action now.

**Your Task**
Your task is to annotate a narrative and identify which parts belong to the story of self, us, 
and now. Note that the narrative will not necessarily follow this order, and may switch 
between these components fluidly, so pay attention to the definitions below and always adhere 
to them when annotating:

**Narrative Code Definitions**
1. Story of Self: A personal experience that reveals the origin of one's values or why they are 
called to leadership. Sometimes called an Origin Story.
2. Story of Us: A communal experience that aims to highlight or generate a sense of shared 
values and experience within a group.
3. Story of Now: An experience that emphasizes the nature of current circumstances and aims 
to motivate an audience to take action to address it.

You will output in a JSON format specified below.
\end{verbatim}

\subsubsection{CoT + Prompt Chaining Prompt 1}
Below \texttt{\{narrative\}} is the full text of the input PN.
\begin{verbatim}
**Public Narrative:**
{narrative}

**Output Instructions**
Step 1. Consider the public narrative above and the definitions of the story of self, story of us, 
and story of now.
Step 2. Identify which sentences comprise the story of self, us, and now.
Step 3. Respond in JSON list for each sentence (don't skip any and do not edit the text), 
indicating a 1 in the respective field if the sentence is part of self/us/now and 0 otherwise. 
Be sure to include ALL lines.
\end{verbatim}

\subsubsection{CoT + Prompt Chaining System Prompt 2}
\begin{verbatim}
Your task is to annotate a public narrative speech according to a specific codebook developed 
by Dr. Marshall Ganz from Harvard. Simply put, Public Narrative says, "Here's who I am, this 
is what we have in common, and here's what we're going to do about it." By mastering the 
practise of crafting a narrative that bridges the self, us, and now, organizers enhance their 
own efficacy and create trust and solidarity with their constituency.

The Public Narrative framework is made up of three components: a Story of Self, a Story of Us, 
and a Story of Now. Each story within this framework follows a fundamental structure that 
grounds abstract values in concrete experiences: Challenge, Choice, and Outcome. Challenge 
shows a moment of adversity or uncertainty. Choice is the decision to respond to the challenge, 
shaped by the storyteller's values. An Outcome demonstrates the result of that choice, 
revealing the stakes and consequences of action (or inaction). For a Story of Self, this 
structure often unfolds through formative moments from one's youth or early leadership 
experiences. In a Story of Us, it highlights collective challenges, an "us's" shared decision, 
and their collective outcome. In a Story of Now, it underscores the present crisis and the 
imperative for action. Together, these components form the structure of the narrative.

**Your Task**
Your task is to annotate a narrative (where the story of self/us/now are already identified) 
to identify which parts denote a challenge, choice, or outcome. Note that the narrative will 
not necessarily follow this order, and may switch between these components fluidly, so pay
attention to the definitions below and always adhere to them when annotating:


**Story Structure Code Definitions**
1. Challenge: A past, present, or potential future situation whose negative impact is to be 
overcome (or has been overcome). May affect either a collective or an individual.
2. Choice: A past, present, or possible future decision to be taken in response to a challenge, 
often reflecting personal values. May be individual or collective.
3. Outcome: A past, present, or possible future result of a choice, whether made or conditional. 
The consequences of this outcome could be known (positive or negative) or unknown, and they may 
affect either a collective or an individual (although more often a collective). 

You will output in a JSON format specified below.
\end{verbatim}

\subsubsection{CoT + Prompt Chaining Prompt 2}
Below, \texttt{\{narrative\}} is the full text of the input PN plus the categorical code annotations output from the first prompt.
\begin{verbatim}
**Public Narrative:**
{narrative}

**Output Instructions**
Step 1. Consider the public narrative above with story of self/us/now annotated, and within 
this context, consider the definitions of challenge, choice, and outcome.
Step 2. Identify which sentences contain a choice, challenge, and/or outcome.
Step 3. Respond in JSON list for each sentence (don't skip any and do not edit the text or 
previous annotations), indicating a 1 in the respective field if the sentence contains a 
challenge/choice/outcome and 0 otherwise. Be sure to include ALL lines.
\end{verbatim}

\section{Final Prompts}\label{appendix:finalprompt}
As mentioned before, the final prompt structure we used was the Chain of Thought + Prompt Chaining. There were three prompts in the chain, where the first was to annotate the categorical codes, the second the structural codes, and the third to annotate the content codes. Each had the same system prompt.

\begin{mdframed}[frametitle={System Prompt:},
    frametitlefont=\bfseries,
    innertopmargin=1em
]
\begin{lstlisting}
<system_prompt>

<role>
You are an experienced qualitative annotator specializing in the Public Narrative Framework developed by Dr. Marshall Ganz. Your exceptional attention to detail enables you to accurately identify and tally specific **CODES** within pieces of provided text.
</role>

<task>
Your task is to annotate a provided textual transcript (a "Public Narrative") according to a predefined codebook. You must identify sections of text that belong to each of three types of **CODES**: **CATEGORICAL CODES**, **STRUCTURAL CODES**, and **CONTENT CODES**. You will perform each coding task independently and in the above sequence.
</task>

<background>
Your task is to annotate a public narrative speech according to a specific codebook developed by Dr. Marshall Ganz from Harvard. Simply put, Public Narrative says, "Here's who I am, this is what we have in common, and here's what we're going to do about it." By mastering the practice of crafting a narrative that bridges the self, us, and now, organizers enhance their own efficacy and create trust and solidarity with their constituency.

The Public Narrative framework is made up of three components: a Story of Self, a Story of Us, and a Story of Now. A Story of Self communicates the values that have called you to leadership; a Story of Us communicates the values shared by those in action; a Story of Now communicates an urgent challenge to those values that demand action now. Each story within this framework follows a fundamental structure that grounds abstract values in concrete experiences: Challenge, Choice, and Outcome. Challenge shows a moment of adversity or uncertainty. Choice is the decision to respond to the challenge, shaped by the storyteller's values. An Outcome demonstrates the result of that choice, revealing the stakes and consequences of action (or inaction). For a Story of Self, this structure often unfolds through formative moments from one's youth or early leadership experiences. In a Story of Us, it highlights collective challenges, an "us's" shared decision, and their collective outcome. In a Story of Now, it underscores the present crisis and the imperative for action. 

Public Narratives also utilize various content markers to help audiences further connect to the stories being told: Story Details, Hope, Values, Vulnerability, and Urgency. Story Details capture the specific moment or experience of the speaker in order to move listeners. Hope moves audiences to positive actions that can emerge from challenges or courageous choices. Values unite audiences and speakers through shared core beliefs. The content marker of Vulnerability reveals the speaker's own authentic experiences in order to increase trust from the audience. Urgency calls for a sense of action and immediacy. Call-to- action encourages listeners to take concreate action. Finally, Dream portrays a desirable past or future state, while Nightmare portrays the consequences of an inadequate response as imagined from the past or the future.

Together, these six components (story of self, story of us, story ofand now; challenge, choice, and outcome) form the structure of the narrative. The other eight content markers help support and enhance these six components and the narrative in general. You are an annotator who is following the public narrative framework detailed above. Your task is to annotate a narrative and identify which parts belong to the story of self, us, and now, which parts denote a challenge, choice, or outcome, and which parts denote story details, hope, values, vulnerability, call-to-action, dream, and nightmare. Note that the narrative will not necessarily follow this order, and may switch between these components fluidly. 
</background>

</system_prompt>

You will output in a JSON format specified below.
\end{lstlisting}
\end{mdframed}

\begin{mdframed}[frametitle={Prompt 1:},
    frametitlefont=\bfseries,
    innertopmargin=1em
]
\begin{lstlisting}
**CATEGORICAL CODES**:
1. Story of Self: Content that reveals formative moments from the storyteller's personal life that illuminate the origins and development of their core values and commitment to action. To do this effectively, the content might include (a) choice points (specific moments of challenge, decision, or transformation that required the storyteller to clarify what mattered most to them); (b) value genesis (explicit connections between particular experiences and the formation of specific values or principles that now guide the storyteller's actions); (c) leadership catalyst (experiences that awakened the storyteller's sense of responsibility or capacity to effect change on issues they care about); (d) authentic struggle (honest portrayal of difficulties, doubts, or failures that shaped the storyteller's understanding and resolve); or (e) developmental arc (a coherent narrative progression showing how the storyteller's identity and purpose evolved through key experiences rather than presenting isolated anecdotes).
2. Story of Us: Content that establishes meaningful collective identity by articulating shared experiences, values, and aspirations that bind the storyteller and audience into a community with common purpose. To do this effectively, the content might include (a) identity markers (specific references to shared characteristics, experiences, or affiliations that define the "us" being addressed); (b) collective memory (references to significant shared historical moments or cultural touchpoints that carry emotional or moral significance for the community); (c) value resonance (explicit articulation of principles that both storyteller and audience recognize as fundamental to their collective identity); (d) challenge recognition (acknowledgment of common struggles or obstacles faced by the community that require collective response); or (e) mutual interdependence (illustrations of how community members' wellbeing and agency are interconnected, establishing shared stake in collective action).
3. Story of Now: Content that creates immediate urgency by presenting a compelling choice point that demands collective action in the present moment to align current reality with shared values. To do this effectively, the content might include (a) critical juncture (framing the present moment as a unique opportunity or pivotal decision point with significant long-term consequences); (b) value-reality gap (highlighting the discrepancy between what the community believes in and current conditions to create productive tension); (c) concrete strategy (outlining specific, feasible steps that can be taken individually and collectively to address the challenge); (d) outcome contrast (vividly portraying both the potential positive future made possible through action and the negative consequences of inaction); or (e) agency activation (explicitly transferring responsibility to the audience by inviting their immediate participation in a clearly defined next step that connects directly to the larger goal).

**Public Narrative:**
{narrative}

**Output Instructions**
Step 1. Consider the public narrative above and the definitions of the story of self, story of us, and story of now.
Step 2. Identify which sentences comprise the story of self, us, and now.
Step 3. Respond in JSON list for each sentence (don't skip any and do not edit the text), indicating a 1 in the respective field if the sentence is part of self/us/now and 0 otherwise. Be sure to include ALL lines.
\end{lstlisting}
\end{mdframed}

\begin{mdframed}[frametitle={Prompt 2:},
    frametitlefont=\bfseries,
    innertopmargin=1em
]
\begin{lstlisting}
**STRUCTURAL CODES**:
1. Challenge: Content that identifies a specific obstacle, problem, or difficult situation that creates tension between current reality and desired values or goals. To do this effectively, the content might include (a) concrete manifestation (specific examples or instances that make abstract problems tangible and immediate); (b) impact articulation (clear description of how the challenge affects individuals or communities in meaningful ways); (c) systemic context (connections between immediate problems and broader patterns or structures that sustain them); (d) emotional resonance (language that captures both the practical and emotional dimensions of the challenge); or (e) value violation (explicit links between the challenge and how it threatens or contradicts core values held by the storyteller and audience).
2. Choice: Content that portrays a significant decision point where values are tested and agency is exercised in response to a challenge. To do this effectively, the content might include (a) option clarity (explicit identification of the different possible responses available at the moment of decision); (b) value tension (illustration of how the choice requires weighing competing priorities or navigating conflicting values); (c) stake recognition (acknowledgment of what stands to be gained or lost through different choices); (d) agency emphasis (focus on the deliberate exercise of power and responsibility in making the choice rather than passive acceptance); or (e) courage dimension (honest portrayal of the fears, risks, or uncertainties that must be faced to make the choice aligned with deeper values).
3. Outcome: Content that describes the consequences-realized or potential-that flow from particular choices in response to challenges. To do this effectively, the content might include (a) tangible results (specific, observable changes that occurred or could occur because of the choice made); (b) learning revelation (insights or understandings gained through experiencing the outcome); (c) value reinforcement (demonstration of how the outcome validates or strengthens commitment to core values); (d) transformative impact (ways in which the outcome changed relationships, perspectives, or circumstances beyond immediate results); or (e) future implication (connections between this outcome and new possibilities, choices, or challenges that emerge as a result).

**Public Narrative:**
{narrative}

**Output Instructions**
Step 1. Consider the public narrative above with story of self/us/now annotated, and within this context, consider the definitions of challenge, choice, and outcome.
Step 2. Identify which sentences contain a choice, challenge, and/or outcome.
Step 3. Respond in JSON list for each sentence (don't skip any and do not edit the text or previous annotations), indicating a 1 in the respective field if the sentence contains a challenge/choice/outcome and 0 otherwise. Be sure to include ALL lines.
\end{lstlisting}
\end{mdframed}

\begin{mdframed}[frametitle={Prompt 3:},
    frametitlefont=\bfseries,
    innertopmargin=1em
]
\begin{lstlisting}
**CONTENT CODES**:
1. Story Details: Content that provides specific, vivid, and sensory elements, intended to create immersive narrative experiences by anchoring abstract concepts in concrete reality. To do this effectively, the content might include (a) sensory information (descriptions that engage multiple senses, allowing audiences to see, hear, smell, taste, or feel aspects of the narrative); (b) concrete particulars (specific people, places, objects, or moments that replace generalizations with precise, memorable imagery); (c) temporal markers (indications of time, sequence, or duration that orient audiences within the chronological flow of events); (d) environmental context (details about physical or social surroundings that establish atmosphere and situate action); or (e) emotional texture (descriptive elements that convey the emotional qualities of an experience rather than merely naming feelings).
2. Hope: Content intended to cultivate a sense of possibility and agency by demonstrating that meaningful change is both necessary and achievable through collective action. To do this effectively, the content might demonstrate (a) balanced realism (acknowledging challenges while identifying viable pathways forward, avoiding both naive optimism and paralyzing despair); (b) historical continuity (connecting concrete examples of past successes to present opportunities, showing that positive change has happened before and can happen again); (c) actionable specificity (identifying concrete, manageable steps that transform overwhelming problems into achievable tasks with visible progress markers); (d) collective efficacy (illustrating how individual contributions gain power when coordinated with others, creating capacity that exceeds the sum of individual efforts); or (e) creative agency (framing uncertainty not as a reason for inaction but as space for intervention and shared authorship of a better future).
3. Values: Content intended, for the audience, to highlight (whether explicitly stated or demonstrated) a core belief, principle, or guiding force that motivates an individual or group of individuals. In the Story of Self, values typically emerge from formative experiences that shaped the storyteller's identity and choices. In the Story of Us, values typically represent the shared principles that bind a community together despite differences. In the Story of Now, values typically create urgency by highlighting the gap between cherished principles and current reality, compelling action to align them.
4. Vulnerability: Content that displays the storyteller's own authentic, meaningful experience(s) in ways that may engender trust or good faith on behalf of the audience. To do this effectively, the content might demonstrate (a) emotional honesty (sharing genuine feelings rather than presenting an idealized image, even when those emotions might be perceived as weakness or imperfection); (b) personal disclosure (revealing private experiences, mistakes, or shortcomings that the storyteller might naturally want to hide); (c) risk-taking (opening oneself to potential judgment or rejection by sharing content that breaks from socially acceptable narratives or exposes one's imperfections; (d) authenticity (presenting oneself genuinely rather than performing a curated version of oneself); or (e) relational transparency (showing the audience the storyteller's true thoughts, including doubts, confusion, or evolving understanding).
5. Urgency: Content that establishes a case for immediate action by highlighting the time-sensitive nature of the challenge and the consequences of delay. To do this effectively, the content might demonstrate (a) temporal significance (explaining why this particular moment presents a unique opportunity or critical juncture that may not persist); (b) escalating stakes (illustrating how delays in addressing the issue will lead to worsening conditions or diminishing options for effective response); (c) moral imperative (framing prompt action as an ethical responsibility that cannot be deferred without compromising core values); (d) opportunity costs (revealing what stands to be lost if action is postponed or what might be gained only through timely intervention); or (e) momentum dynamics (showing how acting now can capitalize on existing energy and resources in ways that become less viable with the passage of time).
6. Call-to-action: To do this effectively, the content might include (a) behavioral specificity (clearly defining what action is being requested with enough detail that audiences know exactly what to do); (b) capacity matching (tailoring the requested action to align with the audience's realistic abilities, resources, and commitment levels); (c) impact transparency (explaining how the requested action connects to broader outcomes and the specific difference it will make); (d) immediate accessibility (providing all necessary information, tools, or pathways needed to take action without significant barriers); or (e) collective framing (positioning individual actions within a broader community effort, emphasizing how personal participation contributes to shared goals and creates belonging).
7. Dream: To do this effectively, the content might include (a) concrete visualization (specific, tangible details that allow audiences to mentally inhabit a better future rather than merely abstractly conceiving it); (b) value embodiment (showing how core principles and commitments would be realized and lived in practice if the desired change were achieved); (c) contrast illumination (highlighting the meaningful differences between current reality and potential future in ways that clarify what's at stake); (d) achievable idealism (balancing aspirational vision with plausible pathways, creating a future that stretches beyond present limitations while remaining within reach of coordinated effort); or (e) personal relevance (connecting the broader vision to individuals' lives, demonstrating how the dream future would positively impact them, their loved ones, and their communities).
8. Nightmare: To do this effectively, the content might include (a) logical extension (projecting current troubling trends forward to their natural conclusion to reveal hidden dangers); (b) experiential proximity (bringing distant or theoretical harms into immediate emotional range through vivid, relatable scenarios); (c) preventable tragedy (emphasizing that negative outcomes are not inevitable but contingent on current choices and actions); (d) vulnerable focus (highlighting impacts on specific people or communities who would bear disproportionate burdens in the nightmare scenario); or (e) moral accountability (framing inaction as an active choice with ethical implications, establishing responsibility for allowing preventable harm to occur).

**Public Narrative:**
{narrative}

**Output Instructions**
Step 1. Consider the public narrative above with story of self/us/now and challenge/choice/outcome annotated, and within this context, consider the definitions of story details, hope, values, and vulnerability.
Step 2. Identify which sentence contain story details, hope, values, and/or vulnerability.
Step 3. Respond in JSON list for each sentence (don't skip any and do not edit the text or previous annotations), indicating a 1 in the respective field if the sentence contains story details/hope/values/vulnerability/urgency/call-to-action/dream/nightmare and 0 otherwise. Be sure to include ALL lines.
\end{lstlisting}
\end{mdframed}

\end{document}